\documentclass{article}

\PassOptionsToPackage{numbers, compress}{natbib}

    \usepackage[final]{neurips_2023}

\usepackage[utf8]{inputenc} 
\usepackage[T1]{fontenc}    
\usepackage{hyperref}       
\usepackage{url}            
\usepackage{booktabs}       
\usepackage{amsfonts}       
\usepackage{nicefrac}       
\usepackage{microtype}      
\usepackage{xcolor}         
\usepackage[english]{babel}
\addto\captionsenglish{
  \renewcommand{\contentsname}%
    {Table of Contents}%
}
\usepackage{enumitem}

\usepackage{amsmath,amsfonts,bm,amsthm,amssymb,mathrsfs,algorithm,algorithmic,url,xcolor,mathtools}
\usepackage{booktabs,enumitem,multicol,caption,subcaption,pifont,tikz,listings,wrapfig,lipsum,circledsteps,arydshln,hyperref,multirow}
\usepackage{relsize,footnote,makecell,graphicx,microtype}

\hypersetup{
    colorlinks,
    linkcolor={blue!50!black},
    citecolor={blue!50!black},
    urlcolor={red!50!black}
}

\theoremstyle{plain}

\theoremstyle{definition}

\theoremstyle{remark}

\newtheorem*{theorem*}{Theorem}

\def\Secref#1{Section~\ref{#1}}

\def\eqref#1{(\ref{#1})}

\def\1{\bm{1}}

\def\vone{{\bm{1}}}

\def\va{{\bm{a}}}

\DeclareMathAlphabet{\mathsfit}{\encodingdefault}{\sfdefault}{m}{sl}
\SetMathAlphabet{\mathsfit}{bold}{\encodingdefault}{\sfdefault}{bx}{n}

\def\gC{{\mathcal{C}}}
\def\gD{{\mathcal{D}}}

\def\gH{{\mathcal{H}}}

\def\gL{{\mathcal{L}}}
\def\gM{{\mathcal{M}}}

\def\gR{{\mathcal{R}}}

\def\gV{{\mathcal{V}}}

\def\gY{{\mathcal{Y}}}

\def\sA{{\mathbb{A}}}

\def\sS{{\mathbb{S}}}

\newcommand{\E}{\mathbb{E}}

\newcommand{\R}{\mathbb{R}}

\newcommand{\cbr}[1]{\left\{#1\right\}}
\newcommand{\br}[1]{\left(#1\right)}
\newcommand{\sbr}[1]{\left[#1\right]}
\newcommand{\given}{\,|\,}

\newcommand{\ie}{\textit{i.e.}}
\newcommand{\eg}{\textit{e.g.}}

\newcommand{\etc}{\textit{etc.}}
\newcommand{\myparagraph}[1]{\textbf{#1~~}}

\newcommand{\dlm}{\mathrm{DLM}}

\definecolor{brickred}{rgb}{0.8, 0.25, 0.33}
\definecolor{darkspringgreen}{rgb}{0.09, 0.45, 0.27}
\definecolor{applegreen}{rgb}{0.55, 0.71, 0.0}
\definecolor{brightmaroon}{rgb}{0.76, 0.13, 0.28}
\definecolor{burgundy}{rgb}{0.5, 0.0, 0.13}

\title{Preference-grounded Token-level Guidance for Language Model Fine-tuning}

\author{
Shentao Yang\textsuperscript{1},~~~~Shujian Zhang\textsuperscript{1},~~~~Congying Xia\textsuperscript{2},~~~~Yihao Feng\textsuperscript{2},\\ \textbf{Caiming Xiong}\textsuperscript{2},~~~~\textbf{Mingyuan Zhou}\textsuperscript{1}
\\
\textsuperscript{1}The University of Texas at Austin~~~~~~~~~~
\textsuperscript{2}Salesforce Research
\\
\texttt{shentao.yang@mccombs.utexas.edu},~~~~
\texttt{yihao.ac@gmail.com},\\
\texttt{mingyuan.zhou@mccombs.utexas.edu}
}

\begin{document}

\maketitle

\begin{abstract}
  Aligning language models (LMs) with preferences is an important problem in natural language generation. A key challenge is that preferences are typically provided at the \textit{sequence level} while LM training and generation both occur at the \textit{token level}.
  There is, therefore, a \textit{granularity mismatch} between the preference and the LM training losses, which may complicate the learning problem.
  In this paper, we address this issue by developing an alternate training process, where we iterate between grounding the sequence-level preference into token-level training guidance, and improving the LM with the learned guidance.  
  For guidance learning, we design a framework that extends the pairwise-preference learning in imitation learning to both variable-length LM generation and the utilization of the preference among multiple generations.
  For LM training, based on the amount of supervised data, we present two \textit{minimalist} learning objectives that utilize the learned guidance.
  In experiments, our method performs competitively on two distinct representative LM tasks --- discrete-prompt generation and text summarization.
  Source codes are released at \href{https://github.com/Shentao-YANG/Preference_Grounded_Guidance}{https://github.com/Shentao-YANG/Preference\_Grounded\_Guidance}.
\end{abstract}

\vspace{-2.2em}
\textcolor{darkspringgreen}{
\section*{Update (01/07/25)}
In our follow-up work \citep{yin2025segmentingtextlearningrewards}, we developed new techniques to successfully scale up the token-level RLHF framework in this paper to PPO + LLMs.
As in this paper, we observed strong gain over the classical bandit RLHF, as tabulated in the following Table~\ref{table:heads_up}.
\begin{table}[H]
\vspace{-1em}
\captionsetup{font=small}
\caption{
Performance comparison between token-level RLHF and bandit RLHF on PPO-trained LM policy, with the 8B-parameter Llama-family backbone model. The judge model is GPT-4o.
For each backbone model, the highest value of each column is in bold.
See Section 4.1 of \citet{yin2025segmentingtextlearningrewards} for experimental details.
} 
\label{table:heads_up}
\centering 
\begin{tabular}{@{}ccccc@{}}
\toprule
Action Space   & Backbone  Model            & AlpacaEval 2 (LC) & Arena-Hard    & MT-Bench      \\ \midrule
\textbf{Token} & Llama-3.1 Instruct 8B & \textbf{45.81}    & \textbf{49.3} & \textbf{7.93} \\
Bandit         & Llama-3.1 Instruct 8B & 40.77             & 36.6          & 7.76          \\ \midrule
\textbf{Token} & Llama-3 SFT 8B        & \textbf{23.84}    & \textbf{26.0} & \textbf{7.13} \\
Bandit         & Llama-3 SFT 8B        & 21.20             & 18.7          & 7.11          \\ \bottomrule
\end{tabular}
\vspace{-1em}
\end{table}
}
\textcolor{darkspringgreen}{
Our follow-up work \citep{yin2025segmentingtextlearningrewards} adopts the framework in this paper and addresses the over-granular issue of token-level RLHF.
Our new proposal is to \textbf{\textit{assign reward to each semantically complete text segment}}, rather than per-token.
Details are in \url{https://arxiv.org/pdf/2501.02790}.
}

\textcolor{darkspringgreen}{
We have also applied the idea of dense reward RLHF  to text-to-image diffusion model with DPO-style explicit-reward-free alignment method \citep{yangdense}; and again observed its significant advantage over the classical DPO loss (bandit RLHF).
More details are in \url{https://arxiv.org/pdf/2402.08265}.
}

\section{Introduction}\label{sec:intro}

Language models (LMs) have been successfully trained with token-level cross-entropy losses, where each token position has a corresponding term in the overall training losses \citep{bengio2000neural,vaswani2017attention,rennie2017self,gpt2018,devlin2018bert,gpt2,bart2019,fan2020bayesian,liu2019roberta,gpt32020,zhang2021bayesian}.
Recent studies have shown that LMs can be further improved 
by aligning them with preferences from human feedback \citep{stiennon2020learning,wu2021recursively,instructgpt2022,gpt42023} or automatic evaluation metrics \citep{caspi2021,fantasticrewards2022,bai2022constitutional}. 
Typically, the preferences are  only provided at the \textit{sequence level}, \eg, ``Which of the two generated text sequences is better?'' 
%
To align LMs with sequence-level preferences, there exist a variety of approaches, 
such as applying external filters to the training texts \citep{xu2020recipes}, performing supervised learning on some curated/improved datasets \citep{hancock2019learning,solaiman2021process,scheurer2022training}, and optimizing the LMs based on a learned sequence-level (pairwise-) preference predictor \citep{instructgpt2022,ziegler2019fine,bai2022training,menick2022teaching}. 

While these approaches have contributed to the development of several revolutionary products \citep[\eg,][]{bai2022constitutional,gpt42023},
a mismatch issue has emerged from the perspective of guiding LM fine-tuning. Concretely, the sequence-level preference is not grounded into the token level, where LM training losses occur. This means that there is a \textit{mismatch}  in granularity between the feedback and training losses --- the preference is coarse-grained while the training losses are fine-grained.
This issue is similar to the delayed-feedback problem in reinforcement learning (RL) \citep{andrychowicz2017hindsight,liu2018competitive,aim2021}, where informative  feedback is available only at the end of the trajectory (sequence) and not at any of the intermediate timesteps.
Previous studies have noted that this problem could have a negative impact on the empirical performance of the resulting LMs \citep{takanobu2019guided,wang2020learning}, as it introduces a more challenging learning problem characterized by higher gradient variance and lower sample efficiency to achieve the learning goal \citep{sqltext2021,snell2022offline}.
To address this granularity mismatch, we focus on the following question:
\textit{How can we effectively ground sequence-level preference into token-level guidance for LM fine-tuning?}
We propose an alternate training process that alternates between two stages: learning preference-grounded token-level guidance and improving the LM using the learned guidance.
This alternate process reduces the requirement on supervised data and targets the low-data regime, \eg, few/zero-shot learning, where task-specific supervised (pre-)training is infeasible and initial LMs have weak zero-shot abilities.

To ground the sequence-level preference into token-level guidance, we propose a framework for learning a token-level ``reward'' function\footnote{We use the words ``guidance'' and ``reward'', ``fine-tuning'' and ``training'' interchangeably, depending on the specific context.}, inspired by reward-learning-from-preferences in the imitation learning  (IL) literature \citep{christiano2017deep,trex2019,drex2020}.
Specifically, we train the token-level rewards such that the corresponding evaluation for a generated text sequence reflects 
the preference among
multiple alternative generations, where the preference comes from task-specific evaluation.
 While \textit{summation} is classically used in IL to aggregate the learned token-level rewards into the text-sequence
evaluation, 
LM tasks can be different from classical IL tasks.
To cater to LM generations,
our 
guidance-learning framework can accommodate more careful choices of the aggregation function beyond the classical summation. For instance, in generating text prompts to steer an LM for text classification, a ``key token'' in the prompt may be more effective than several mediocre tokens. Hence, using \textit{maximum} to aggregate the token-level rewards may better reflect the text-sequence quality than \textit{summation}. 



To utilize the learned preference-grounded guidance in LM training, we present two \textit{minimalist} learning objectives that contain only a minimal number of hyperparameters. These two objectives respectively target different amounts of supervised data in the specific LM task. 
We evaluate our framework on two distinct representative LM tasks: generating discrete text prompts for few-shot text classification and text summarization. 
On both tasks, our method exhibits competitive performance.



\section{Main Method}\label{sec:main_method}


Before diving into technical details, we will first establish the notation, provide some background on classical pairwise-preference learning, and offer an overview of our preference-grounding process.

\myparagraph{Notation.}\label{sec:background}
In most LM tasks, we are given a dataset $\gD = \{(x^{i}, y^{i})\}_{i=1}^N$ of $N$ supervised examples, where $x$ is the input to the LM, which can be a dummy, 
and $y$ is the target text-sequence.
We denote the LM parameterized by $\theta$ as $\pi_\theta$.
The $t^{\text{th}}$ generated token is denoted as $a_t$, given by $a_t \sim \pi_\theta(\cdot \given s_t)$, where the context for token generation at step $t\geq 0$ is denoted as $s_t$, consisting of the LM input $x$ and the previously generated tokens $a_{< t} = (a_0, \ldots, a_{t-1})$. 
Specifically, $s_0 = x$ and $\forall\, t > 0, s_t = (x, a_{< t})$. 
The full generated text-sequence of length $T$ is denoted as $\va = (a_0, \ldots, a_{T-1})$.
In most LM tasks, we have a task-specific evaluation metric $\gR(s_T, y) \in \R$ that depends on the final context $s_T$ of the generated sequence and the target sequence $y$, with $s_T = (x, \va)$.
The objective of LM training is often to maximize the expected task-specific evaluation, which can be expressed as
\begin{equation*}
\textstyle
        \max_\theta \E_{(x,y)\sim \mathcal D}
        \E_{\va \sim \prod_{t=0}^{T-1} \pi_\theta(a_t \given s_t)}\sbr{\gR(s_T = (x, \va), y)}.
\end{equation*}
We model the learned token-level guidance as a bounded (reward) function $r_\phi( s_t, a_t ) \in [0,1]$, parametrized by $\phi$. 
Unlike the original sequence-level preference or evaluation that is only available at the final step $T$, the trained $r_\phi$ can  
densely guide the token selection at each 
step $t \leq T$.

\myparagraph{Pairwise Preference Learning.}
In reward-learning-from-preferences \citep[\eg,][]{christiano2017deep,trex2019,drex2020}, a dense reward function is learned such that the \textit{sum-aggregated} reward for the entire generation trajectory aligns with the pairwise preference between two trajectories.
In the context of LM generation, suppose we have two text-generation trajectories $\tau^i$ and $\tau^j$ associated with the same LM input and target $(x,y)$, taking the form $\tau^i = \{(s^i_0, a^i_0), \ldots, (s^i_{T^i-1}, a^i_{T^i-1})\}$ with sequence lengths $T^i$ and $T^j$, respectively.
Assume that $\tau^i$ is preferred over $\tau^j$, denoted as $\tau^i \succ \tau^j$. 
A token-level reward function $r_\phi(s_t, a_t)$ is learned by requiring $\sum_{t=0}^{T^i-1} r_\phi(s^i_t, a^i_t) > \sum_{t=0}^{T^j-1} r_\phi(s^j_t, a^j_t)$.
Following the  Bradley-Terry model of preferences \citep{bradley1952rank}, the pairwise-preference loss for reward-function learning is
\begin{equation}\label{eq:original_pref_rew}\textstyle
\resizebox{0.67\textwidth}{!}{%
    $
    \ell(\phi) = -\log \sbr{ \exp\br{\sum_{t=0}^{T^i-1}r_\phi(s^i_t, a^i_t) } \Big/ \sum_{k \in \{i, j\}}\exp\br{\sum_{t=0}^{T^k-1}r_\phi(s^k_t, a^k_t)} } \,,
$%
}
\end{equation}
which is often interpreted as binary classification in the literature \citep{ranksvm1999,rankingboosting2003,ranknet2005}.
In Eq.~\eqref{eq:original_pref_rew}, \textit{summation} $\sum (\cdot)$ is used to aggregate the learned token-level rewards into a parametrized sequence-level evaluation.


\myparagraph{Overview.} 
To ground the \textit{sequence-level} preference into \textit{token-level} guidance for LM training and thereby address the granularity mismatch discussed in \Secref{sec:intro},
we present an alternate learning process that alternately learns the token-level guidance and trains the LM using the learned guidance. 
 
\begin{figure}[t]
\centering
\includegraphics[width=.85\textwidth]{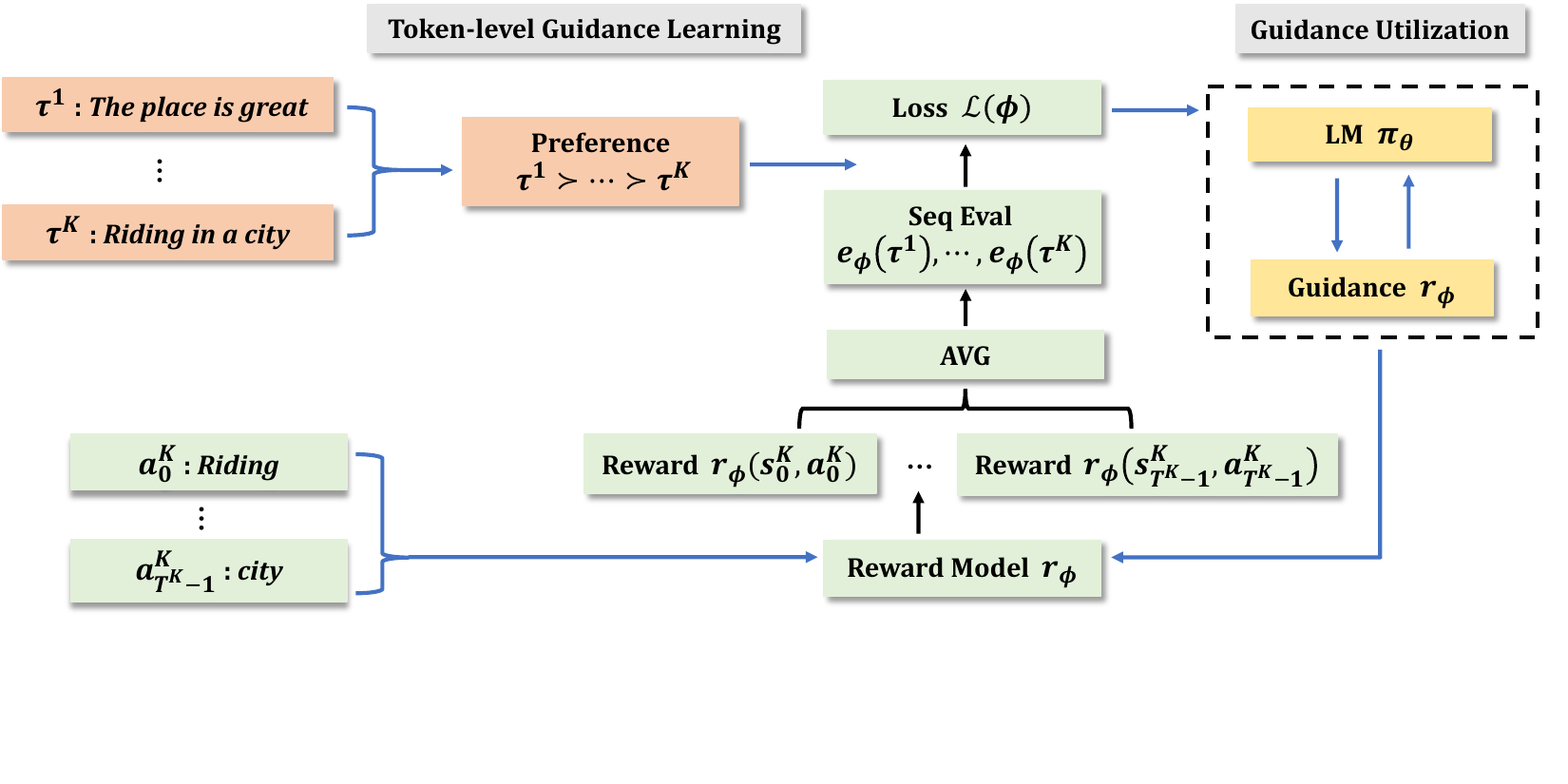} 
\vspace{-3.4em}
\captionsetup{font=small}
\caption{Overview of the proposed framework. 
``AVG'' denotes \textit{average}, which is an example of the aggregation function $f(\cdot)$ discussed in \Secref{sec:method_agg_func}. 
``Seq Eval'' refers to the parametrized sequence-level evaluations. 
The model choice of the reward function and  LM depends on the specific task and is discussed in \Secref{sec:exp}.
}
\label{fig:pipeline}
\end{figure}

For learning the preference-grounded guidance, in \Secref{sec:method_framework} we propose a framework that learns a token-level reward function that reflects the preference among multiple generated sequences. 
To utilize the learned preference-grounded guidance, 
based on the amount 
of  supervised  data in the specific task, 
in \Secref{sec:policy_training} we present two \textit{minimalist} LM training approaches  that require only minimal tuning. 
In our framework, we iterate between the above two steps to mitigate the 
distribution shift between the text sequences used to train the reward function and the text sequences evaluated by the reward function during LM training, taking into account that LMs can evolve during the training process.
Our alternate-learning procedure is illustrated in Fig.~\ref{fig:pipeline}. 

\subsection{Token-level Guidance Learning for Preference Grounding}\label{sec:method_framework}

Instead of applying the pairwise approach discussed above, 
we utilize the preference among multiple generated text sequences to learn the reward function $r_{\phi}(s_t ,a_t)$. 
Intuitively, we use more information to train the reward function at each optimization step. 
Therefore, our approach can be more efficient and effective, especially when the optimization budget is limited.

Concretely, suppose we have $K\geq 2$ generated text-sequences $(\va^1, \ldots, \va^K)$ for the same LM input and target $(x,y)$, with the associated generation trajectories $(\tau^1, \ldots, \tau^K)$ and with possibly unequal sequence-lengths $(T^1, \ldots, T^K)$.
Assume that there is a preference ordering among these $K$ sequences, where by ``preference'' we mean a ranking of text sequences based on some evaluations of full text-sequences.
For description simplicity, let the preference ordering be $\va^1 \succ \dots \succ \va^K \iff \tau^1 \succ \dots \succ \tau^K$. 
We make no assumption about the source of  preference. 
It may come from human ranking 
or some task-specific evaluation metric $\gR$ on the full text-sequences, \eg, the descending ordering of the text-sequence evaluations $\gR(s^1_{T^1}, y) > \cdots > \gR(s^K_{T^K}, y)$.

For a trajectory $\tau^k = \{(s^k_0, a^k_0), \ldots, (s^k_{T^k-1}, a^k_{T^k-1})\}$, our desired token-level reward function $r_\phi$ generates a reward for each step as $\{r_\phi(s^k_t, a^k_t)\}_{t=0}^{T^k-1}$.
A sequence-level evaluation $e_\phi(\tau^k)$ for trajectory $\tau^k$ can be obtained by 
$e_\phi(\tau^k) = f(\{r_\phi(s^k_t, a^k_t)\}_{t=0}^{T^k-1})$, 
where $f(\cdot)$ is the aggregation function over all per-step rewards, \eg, the classical \textit{summation} $\sum(\cdot)$.
Our goal is to train $r_\phi$ such that these parametrized sequence-level evaluations $\{e_\phi(\tau^k)\}_{k=1}^K$ align with the given preference ordering $\tau^1 \succ \dots \succ \tau^K$.
Through this, the sequence-level preference is grounded into token-level rewards $r_\phi$.

Under the Plackett--Luce choice model \citep{plackett1975analysis,luce2012individual}, the parametrized sequence evaluations $\{e_\phi(\tau^k)\}_{k=1}^K$ induce a probability distribution over all possible permutations of the integers $\{1, \ldots, K\}$.
We want to maximize the likelihood of the given preference ordering $\mathrm{ord} = (1,\ldots,K)$, \ie,
\begin{equation} \label{eq:listmle} \textstyle
\resizebox{.94\linewidth}{!}{%
$
     \begin{aligned}
         \min_\phi \gL(\phi) =: -\log P\br{\mathrm{ord} \given \{e_\phi(\tau^k)\}_{k=1}^K},  \;
         P\br{\mathrm{ord} \given \{e_\phi(\tau^k)\}_{k=1}^K} = \prod_{{k=1}}^{K} \cbr{\exp(e_\phi(\tau^k)) \bigg/ \sum_{{i=k}}^{K}\exp(e_\phi(\tau^i))}.
     \end{aligned}
$%
}
\end{equation}
When $K=2$ and $f(\cdot)$ denotes \textit{summation}, Eq.~\eqref{eq:listmle} reduces to the classical pairwise-preference loss in Eq.~\eqref{eq:original_pref_rew}.
Therefore, our reward-learning loss can be viewed as an extension of the classical pairwise loss.
Further, Eq.~\eqref{eq:listmle} extends the ListMLE loss \citep{listmle2008} in recommender systems into preference learning under multiple 
variable-length trajectories. 

Algo.~\ref{algo:reward_learning} summarizes our reward-learning framework by describing an online-learning routine for training $r_\phi$. 
An offline or hybrid version can be obtained with minor changes. 

\setlength{\textfloatsep}{0.1cm}
\setlength{\floatsep}{0.1cm}
\begin{algorithm}[tb]
\caption{A learning routine for the preference-grounded token-level reward function $r_\phi$. }
\label{algo:reward_learning}
\begin{algorithmic}
\STATE \textbf{Input:} The LM $\pi_\theta$, initialized reward $r_\phi$, aggregation function $f(\cdot)$, reward-training steps $M_{\mathrm{rew}}$.
\FOR{$\mathrm{iter} \in \{1, \ldots,M_{\mathrm{rew}}\}$}
\STATE Use $\pi_\theta$ to generate $K$ sequences $\{\va^k\}_{k=1}^K$; and get the preference ordering among $\{\va^k\}_{k=1}^K$.
\STATE With $f(\cdot)$, get sequence evaluations $\{e_\phi(\tau^k)\}_{k=1}^K$ from $r_\phi$; and optimize $r_\phi$ by Eq.~\eqref{eq:listmle}.
\ENDFOR
\end{algorithmic}
\end{algorithm}
\setlength{\textfloatsep}{0.2cm}
\setlength{\floatsep}{0.1cm}


\myparagraph{The Choice of Aggregation Function $f(\cdot)$.}  \label{sec:method_agg_func}
In classical IL tasks such as robotics \citep{todorov2012mujoco}, the robots are trained to stand or walk as long as possible.
In this scenario, \textit{summation} is  a natural choice for the aggregation function $f(\cdot)$.
However, in many text generation tasks,  
such as summarization,
the generation quality may not be directly associated with the length of the generated text sequence. 
Nevertheless, suppose the token-level rewards are positive (\ie, $r_\phi > 0$), a longer sequence naturally has a higher sum of per-step rewards than a shorter one, which can bias $r_\phi$ towards automatically ranking longer sequences higher.
This bias can hinder our reward-learning goal of aligning $\{e_\phi(\tau^k)\}_{k=1}^K$ with the given preference ordering.
A na\"ive numeric example is additionally provided in Appendix~\ref{sec:avg_agg}.

To mitigate the potential length bias in the classical \textit{summation}, 
we discuss three alternative choices of the aggregation function $f(\cdot)$:  \textit{average}, \textit{soft maximum}, and \textit{soft minimum}.

\myparagraph{\textit{Average}.}
We define the \textit{average-aggregated} sequence-level evaluation $e^{\mathrm{avg}}_\phi(\tau^k)$ for trajectory $\tau^k$ as
\begin{equation} \label{eq:avg_agg}\textstyle
e^{\mathrm{avg}}_\phi(\tau^k) = \frac{C}{T^k}\sum_{t=0}^{T^k-1} r_\phi(s^k_t, a^k_t), \quad C = \frac{1}{K} \sum_{k=1}^{K} T^k,
\end{equation}
where $C$ is the average length of the $K$ sequences.
Multiplied by the average length $C$ has the benefit of scaling $e^{\mathrm{avg}}_\phi$ to the scale of $e^{\mathrm{sum}}_\phi$, which ensures numerical-scale consistency with $e^{\mathrm{sum}}_\phi$ and thus reduces hyperparameter tuning when switching from \textit{summation} to \textit{average} aggregation.

\myparagraph{\textit{Soft Maximum}.}
We define the \textit{soft-maximum-aggregated} sequence-level evaluation $e^{\max}_\phi(\tau^k)$ as
\begin{equation} \label{eq:max_agg}\textstyle
e^{\max}_\phi(\tau^k) = C \times \beta \cdot \log \sbr{\sum_{t=0}^{T^k-1} \exp(r_\phi(s^k_t, a^k_t)  / \beta)},
\end{equation}
where $C$ is the average trajectory-length in Eq.~\eqref{eq:avg_agg} and $\beta$ is the temperature parameter.

\myparagraph{\textit{Soft Minimum}.} The \textit{soft-minimum-aggregated} sequence-level evaluation $e^{\min}_\phi(\tau^k)$ 
follows Eq.~\eqref{eq:max_agg} except for changing $\beta$ to $-\beta$.

\subsection{LM Training with Preference-grounded Token-level Guidance} \label{sec:policy_training}

Considering the supervised-data availability, we present two \textit{minimalist} LM training objectives that utilize the learned preference-grounded guidance:
\textbf{1)} a REINFORCE-style update when there is no supervised data; 
\textbf{2)} reward-weighted MLE when there are sufficient data.
Our LM training directly starts from raw pre-trained checkpoints, without task-specific supervised pre-training.
This choice is to keep the algorithm general and consistent in both situations we consider, since task-specific pre-training may not be feasible in the setting of few/zero-shot learning.

As shown in Algo.~\ref{algo:reward_learning}, we train the reward function $r_\phi$ by the sequences sampled from LM~$\pi_\theta$.
Since task-specific pre-training to $\pi_\theta$ is not assumed, over the course of training, $\pi_\theta$ itself can evolve from a less-preferred distribution to a highly-preferred one.
To mitigate the impact of this distribution shift and keep $r_\phi$ as accurate guidance for LM training, we periodically re-estimate $r_\phi$ during the first half of the LM-training process\footnote{
In our preliminary study, we observed that this choice (``retraining the reward model only during the \textit{first half} of the LM-training process'') can save about 25-30\% compute without hurting the performance much, compared to the vanilla reward retraining (``retraining the reward model throughout the \textit{entire} LM-training process'').
}, motivated by recent works in model-based RL \citep{mnm2021,wmopo2021,jointmatching2022,sdmgan2022,wmbrl2022}.  

\myparagraph{Without Supervised Data.} \label{sec:policy_training_without_data}
When the LM $\pi_\theta$ needs to discover good text generations by itself, the learned token-level reward $r_\phi$ can be used to provide dense guidance on generating each token, \ie, given the generation context $s_t$, select the next token $a_t$ such that $r_\phi(s_t, a_t)$ is high.
Intuitively, for a generation trajectory $\tau$, if $\forall\, (s_t, a_t) \in \tau, r_\phi(s_t, a_t)$ is high, then the corresponding sequence-level evaluation $e_\phi(\tau) = f(\{r_\phi(s_t, a_t)\}_{t=0}^{T-1})$ can be also high, \eg, the average or summation of token-level rewards.
The associated text sequence $\va$ will thus be preferable since $r_\phi$ is trained to reflect the sequence-level preference (\Secref{sec:method_framework}).
Through $r_\phi$, the sequence-level preference is grounded into dense token-level guidance for LM training, without granularity mismatch or feedback delay.

With the learned $r_\phi$, a \textit{minimalist} implementation of this LM-training idea is the discrete-variable optimization problem 
\begin{equation*} \textstyle
\max_\theta \E_{t\sim \text{Uniform}\{0, \ldots, T-1\}}\E_{a_t \sim \pi_\theta(\cdot \given s_t)}[r_\phi(s_t, a_t)] \,,
\end{equation*}
 for each timestep $t$ of which we calculate its gradient by the classical REINFORCE method \citep{glynn1990likelihood,reinforce1992,fu2006gradient} since it can cope with a large vocabulary size. 
 Here, $T$ denotes a generic sequence length.
Additionally, since we want multiple text generations in typical LM tasks, instead of only one, we relax the convergence of the REINFORCE method by adding a standard max-entropy gradient, which can help capture multiple good behavior-modes \citep{rlenergypolicy2017,sac2018,sacnew2018}.
Thus, the LM $\pi_\theta$ is trained by the gradient
\begin{equation} \label{eq:prompt_pol_objective} \textstyle
        \E_{t\sim \text{Uniform}\{0,\ldots,T-1\}}\left\{ \E_{  a_t \sim \pi_\theta(\cdot \given s_t)}[r_\phi(s_t, a_t) \cdot \nabla_{\theta} \log \pi_{\theta}(a_t \given s_t)] 
           + \alpha \cdot \nabla_\theta \gH(\pi_\theta(\cdot \given s_t))\right\} \,,
\end{equation}
where $\gH(\pi_\theta(\cdot \given s_t))$ is the Shannon entropy of $\pi_\theta(\cdot \given s_t)$ and $\alpha$ is a balancing coefficient.

\myparagraph{With Supervised Data.} \label{sec:policy_training_with_data}
With a labelled dataset $\gD = \{(x^{i}, y^{i})\}_{i=1}^N$ 
and with the learned preference-grounded guidance $r_\phi$, a \textit{minimalist} enhancement of the classical MLE LM-training is the token-level weighted-MLE, where the per-token weight is given by the learned reward-function $r_\phi$.
Our intention is to emphasize the important tokens in the given sequence $y$ and downweight the unimportant ones, where the token importance given by $r_\phi$ grounds the sequence-level preference.
Intuitively, this weighting scheme can better utilize the LM capacity and the optimization budget, and may thus improve upon the vanilla supervised loss \citep{crr2020,caspi2021}.
Specifically, the LM $\pi_\theta$ is trained~by 
\begin{equation} \label{eq:sum_pol_objective} \textstyle
\resizebox{0.94\textwidth}{!}{%
    $
          \min_{\theta} -\E_{(x,y) \sim \gD}\sbr{\sum_{t=0}^{|y|-1} w_t \cdot \log \pi_\theta(y_t \given s_t)} \,,  
          \text{with }  s_t = (x, y_{< t}) \text{ and } w_t = \frac{r_\phi(s_t, y_t)}{ \sum_{t'=0}^{|y|-1} r_\phi(s_{t'}, y_{t'})}  \,,
     $%
}
\end{equation}
where $|y|$ is the length of the target sequence $y$ and $w_t$ is the self-normalized token-level reward.
The standard self-normalization is used to reduce the gradient variance among the samples in $\gD$.

Algo.~\ref{algo:pol_reward_learning} sums up the entire alternate-learning process,
with the reward-learning routine in Algo.~\ref{algo:reward_learning}.

\setlength{\textfloatsep}{0.1cm}
\setlength{\floatsep}{0.1cm}
\begin{algorithm}[tb]
\caption{An alternate-learning process for the reward function $r_\phi$ and the LM $\pi_\theta$.}
\label{algo:pol_reward_learning}
\begin{algorithmic}
\STATE \textbf{Input:} The dataset $\gD$, initialized LM $\pi_\theta$, initialized reward function $r_\phi$, LM-training steps $M_{\mathrm{LM}}$, reward-retrain period $M_{\mathrm{re}}$, all inputs for training the reward function  specified in Algo.~\ref{algo:reward_learning}.
\STATE \textbf{Initialize} $r_\phi$ by Algo.~\ref{algo:reward_learning}.
\FOR{$\mathrm{iter} \in \{1, \ldots,M_{\mathrm{LM}}\}$}
\IF{$\mathrm{iter} \leq M_{\mathrm{LM}} / 2~\mathrm{and}~\mathrm{iter}~\%~M_{\mathrm{re}}$ == 0}
\STATE Re-train $r_\phi$ by Algo.~\ref{algo:reward_learning} without re-initialization.
\ENDIF
\STATE Optimize $\pi_\theta$ by Eq.~\eqref{eq:prompt_pol_objective} or Eq.~\eqref{eq:sum_pol_objective} with $\gD$ and $r_\phi$.
\ENDFOR
\end{algorithmic}
\end{algorithm}
\setlength{\textfloatsep}{0.2cm}
\setlength{\floatsep}{0.1cm}

\section{Related Work} \label{sec:related_work}


\myparagraph{Guiding Signals for LM Training.} 
One string of works in LM training directly optimizes the LMs against the native sequence-level feedback such as the test-time metric \citep[\eg,][]{rennie2017self,ryang2012framework,ranzato2015sequence,paulus2017deep,shu2021reward,quark2022,snell2022offline}. 
This choice, however, may directly suffer from the delayed-feedback issue discussed in \Secref{sec:intro} and the subsequent high gradient variance and low sample efficiency \citep{sqltext2021,snell2022offline}.
%
%
%
In the recent trend of RL-based LM training, it has been common to incorporate 
a token-level KL penalty towards the uniform distribution \citep{sqltext2021,rlprompt2022}, the initial LM \citep{ziegler2019fine,nlpo2022}, the supervised-fine-tuned model \citep{stiennon2020learning,offlinerldialog2019,jaques2020human,zhang2022passage,instructgpt2022}, or the base momentum model \citep{castricato2022robust}, to add to the delayed/ungrounded feedback. 
Although that KL penalty does impact the RL-based LM training at the token level, it is not tailored to the concrete task or the desired sequence-level feedback.
When combined with the delayed-feedback issue, it could distract the LM training from improving the received feedback/evaluation, especially at the beginning of the text-sequence generation, which can however affect all subsequent token selections.
By contrast, as seen in Eq.~\eqref{eq:prompt_pol_objective}, even when added a max-entropy gradient, 
our preference-grounded token-level guidance can still provide dense, task-specific, and feedback-oriented guidance on the selection of each token.
For a more detailed discussion on the RL formulation of LM generation, the delayed-feedback issue in RL-based LM training, and delayed feedback with KL penalty, please refer to Appendix~\ref{sec:details_rl_nlp}.

In some relatively ``ideal'' settings, prior works have attempted to learn task-specific token-level guidance for LM training. For instance, \citet{shi2018toward} use inverse RL,  \citet{leakgan2018} propose a hierarchical approach, and \citet{yang2018unsupervised} learn LM discriminators; but these methods require abundant expert data for supervised (pre-)training, making them infeasible for the few/zero-shot settings we consider. 
Under the same requirement of sufficient expert data, \citet{adversarialranking2017} learn a sequence-level adversarial-ranking reward and \citet{yu2017seqgan} train a GAN structure. 
They both use Monte-Carlo rollouts to simulate intermediate rewards, which can be computationally expensive and have high variance.  
\citet{le2022coderl} use some values related to the sequence evaluation without explicitly learning per-token rewards.
\citet{rewardgaming2022} learn a token-level error predictor for machine translation, but they rely on expert error-span annotations for each translation, which is highly demanding.

By contrast,
we propose a versatile framework for learning task-specific token-level guidance for LM training that can ground the sequence-level preference. Our approach is not limited to standard LM tasks and is also suitable for the low-data regime, with few assumptions about expert-data availability or preference source. In our experiments, 
we compare our method to recent RL-based approaches that train LM under delayed/ungrounded feedback with KL penalty. We discuss additional related works on prompt generation, text summarization, and aligning LMs with preferences in Appendix~\ref{sec:more_related}.

\section{Experiments}\label{sec:exp}

\label{sec:exp_main} 


We test our framework on two distinct representative  text-sequence generation tasks:
\textbf{1)} input-agnostic discrete text-prompt generation for few-shot text-classification (\Secref{sec:app_prompt}),
\textbf{2)} the classical text summarization (\Secref{sec:app_summa}).
Our LM training directly starts from raw pre-trained checkpoints from HuggingFace~\citep{huggingface2019}, without task-specific supervised pre-training.
Depending on the LM $\pi_\theta$ used in the specific task, our reward function $r_\phi$ can be implemented as either a decoder-only  or an encoder-decoder model.
Similar to prior works \citep[\eg,][]{snell2022offline,menick2022teaching}, given a text sequence $\va$ and an LM input $x$, the causal mask in transformers enables us to get the learned guidance $r_\phi(s_t, a_t)$ at each step of the sequence in parallel.        
Source codes have been publicly  \href{https://github.com/Shentao-YANG/Preference_Grounded_Guidance}{released}.

\subsection{Input-agnostic Discrete-prompt Generation} \label{sec:app_prompt}

\myparagraph{Overview.}
In discrete text-prompt generation \citep[\eg,][]{gpt32020,an2022input}, we input a discrete text-prompt $\va$ and an observation sequence $o$ to a large pre-trained downstream LM $\pi_\dlm(\cdot \given \va, o)$ to directly classify text~$o$, without finetuning $\pi_\dlm$.
We follow the classical setting \citep[\eg,][]{schick2020s,rlprompt2022} to perform classification by selecting tokens corresponding to some predefined class labels.
In our input-agnostic setting, the generated prompt is independent of the observation $o$. 
During inference time, only the learned prompts are used and the LM $\pi_\theta$ is discarded.
The initial input $x$ to $\pi_\theta$ is a dummy, and the target~$y$ is the class label.
We also adopt the standard few-shot setting \citep{perez2021true}, where both the training and validation sets have $16\, (o, y)$-pairs per class.
With a fixed length $T$, the goal is to find discrete text-prompts $\va = (a_0, \ldots, a_{T-1})$ that have high test accuracy.
We simulate the sequence-level preference by the stepwise metric in \citet{rlprompt2022}, \ie, the higher value the better prompt.
This choice ensures a fair comparison and avoids a potential overfitting --- training and testing the LM on the same evaluation metric ``accuracy''.
Appendix~\ref{sec:details_prompt_tasks} discusses more details about the prompt task.

\myparagraph{LM Training, Implementation, and Datasets.}
Since the prompt-generation task does not assume the availability of supervised data --- the ground-truth prompts, the LM $\pi_\theta$ is trained by the REINFORCE-style update in \Secref{sec:policy_training_without_data} to discover highly-accurate prompts by itself.
We implement our framework on the codebase of RLPrompt \citep{rlprompt2022}, and adopt the standard datasets and most hyperparameter settings in it.
Reward training is reconducted every $1000$ steps during the first $6000$ steps of the LM training process
and has early stopping.
Reward function is learned with $5$ sampled sequences and the temperature in Eq.~\eqref{eq:max_agg} is set as $\beta=2$.
The coefficient $\alpha$ in Eq.~\eqref{eq:prompt_pol_objective} is $\alpha = 2^{-3}$.
Appendix~\ref{sec:more_abla} discusses the choices of these hyperparameters.
The length of the generated prompts is fixed at $5$.
We test on three popular few-shot datasets in prior work \citep[\eg,][]{gao2020making,sun2022black}: two sentiment binary-classification datasets SST-2 \citep{socher2013recursive,zhang2021alignment} and Yelp Polarity \citep{zhang2015character}, and a topic four-way-classification dataset AG News \citep{zhang2015character,zhang2022allsh}.
Additional details on the experiment and datasets are provided in Appendix~\ref{sec:prompt_details}.

\begin{table}[tb]
\captionsetup{font=footnotesize}
\caption{
\footnotesize
Test accuracy on the prompt task.
Best overall result 
is bold and best discrete-prompt result is underlined if different.
The reported results are mean (standard deviation).
We denote 
``BB Tuning-$50$'' for Black-Box Tuning with mixed discrete and soft prompts that tunes the $50$ soft tokens; and ``AVG'', ``SUM'', ``MIN'', ``MAX'' for our method with aggregation function average, summation, soft minimum, and soft maximum (\Secref{sec:method_agg_func}). 
} 
\label{table:prompt_main}
\centering 
\setlength\tabcolsep{20pt}
\resizebox{.93\textwidth}{!}
{
\footnotesize
\begin{tabular}{@{}lllll@{}}
\toprule
\multicolumn{1}{c}{}          & \multicolumn{1}{c}{}                       & \multicolumn{1}{c}{SST-2} & \multicolumn{1}{c}{Yelp P.}       & \multicolumn{1}{c}{AG News}    \\ \midrule
Finetuning                    & Few-shot Finetuning                        & 80.6 {\scriptsize (3.9) }  & 88.7 {\scriptsize (4.7)}  & \textbf{84.9} {\scriptsize (3.6) } \\ \midrule
\multirow{4}{*}{\makecell[l]{Continuous \\Prompt}} & Soft Prompt Tuning \citep{lester2021power}                         & 73.8 {\scriptsize (10.9) } &  88.6 {\scriptsize (2.1)} & 82.6 {\scriptsize (0.9) } \\
                              & BB Tuning-$50$ \citep{sun2022black}  & 89.1 {\scriptsize (0.9)} & 93.2 {\scriptsize (0.5)}  & 83.5 {\scriptsize (0.9) } \\
                              & AutoPrompt \citep{shin2020autoprompt}                                 & 75.0 {\scriptsize (7.6) } & 79.8 {\scriptsize (8.3)}  & 65.7 {\scriptsize (1.9) } \\ \midrule
\multirow{8}{*}{\makecell[l]{Discrete \\Prompt}} & Manual Prompt  \citep{schick2020exploiting}                            & 82.8  & 83.0       & 76.9       \\
                              & In-Context Demo      \citep{gpt32020}                      & 85.9 {\scriptsize (0.7) } & 89.6 {\scriptsize (0.4)}  & 74.9 {\scriptsize (0.8) } \\
                              & Instructions \citep{mishra2021cross}                              & 89.0    & 84.4    & 54.8       \\
                              & GrIPS \citep{prasad2022grips}                                      & 87.1 {\scriptsize (1.5) } & 88.2 {\scriptsize (0.1)}  & 65.4 {\scriptsize (9.8) } \\
                              & RLPrompt    \citep{rlprompt2022}                               & 90.5 {\scriptsize (1.5) } &  94.2 {\scriptsize (0.7)}  & 79.7 {\scriptsize (2.1) } \\ \cmidrule(l){2-5} 
                              & Ours  (AVG / SUM)                                  & \textbf{92.6} {\scriptsize (1.7) } & 94.7 {\scriptsize (0.6)}  &  82.8 {\scriptsize (1.5) } \\
                              & Ours  (MIN)                                  & 91.9 {\scriptsize (1.8) } & 94.4 {\scriptsize (0.8)}  & 82.4 {\scriptsize (1.1) } \\
                              & Ours  (MAX)                                  & 91.2 {\scriptsize (2.5) } & \textbf{94.8} {\scriptsize (0.5)}  & \underline{83.3} {\scriptsize (1.4) } \\ \bottomrule
\end{tabular}
}
\end{table}
\myparagraph{Results.} \label{sec:main_result_prompt}
We compare three variants of our framework with finetuning and with baselines in discrete- and continuous-prompt generation.
Since the generated prompts all have length $5$, in this task, the \textit{average} aggregation is equivalent to \textit{summation}. 
Table~\ref{table:prompt_main} shows the test accuracy, where we rerun the codebase of RLPrompt \citep{rlprompt2022} under the same random seeds and evaluation script as our method.\footnote{There are small discrepancies between our reported RLPrompt results  and the original paper's. We have confirmed our reproduced results both with RLPrompt's authors and with the recent TEMPERA paper \citep{tempera2022}.}
Other baseline results are from the literature \citep{rlprompt2022,tempera2022}.

On all three tested datasets, our method shows competitive and stable results against the strong baselines not only in discrete-prompt generation, but also in heavier continuous-prompt tuning and finetuning the large downstream LM.
Based on \Secref{sec:related_work}, the performance improvement achieved by our method compared to RLPrompt suggests that utilizing the token-level guidance learned by our approach, which grounds the task-specific preference, can be more effective than learning under delayed/ungrounded feedback with KL penalty.
Further, on both Yelp P. and AG News, using MAX aggregation is better than the classical \textit{summation}.
Table~\ref{table:prompt_example} in Appendix~\ref{sec:add_results} shows examples of good generated prompts and their test accuracy.
For instance,
high-quality prompts on the AG News dataset often contain a topic classification keyword, such as ``Tags'' and ``Category''. This aligns with our intuition that good prompts may be identified by a (few) ``key'' token(s), as discussed in Sections \ref{sec:intro} and \ref{sec:method_agg_func}. Thus, the \textit{(soft-)maximum} aggregation may better reflect prompt quality than \textit{summation}.

\subsection{Text Summarization} \label{sec:app_summa}
\myparagraph{Overview.}
In the summarization task, we follow the standard setting \citep[\eg,][]{t52020,nlpo2022}, where a set of supervised samples is available.
The LM input $x$ is the text to be summarized and the target $y$ is the given summary.
We simulate the sequence-level preference by the classical Meteor score \citep{meteor2005} and report the standard ROUGE scores \citep{rouge2004}, to avoid overfitting evaluation metrics as in the prompt task.

\myparagraph{LM Training, Implementation, and Datasets.}
Since a supervised dataset $\gD$ is available in this task, the LM $\pi_\theta$ can be trained by the weighted-MLE objective in \Secref{sec:policy_training_with_data}.
This objective could be more stable and computationally efficient than REINFORCE-style methods in tasks of long-sequence generation.
Due to limited computing resources, unless explicitly mentioned,
we use the standard T5-small model \citep{t52020} for both the LM and reward function.
The reward training is simply $1$ epoch of training on randomly sampled $10\%$ of the training set and is repeated every $0.5$ epochs during the first $2$ epochs of LM training. 
Reward function is learned with $3$ sampled sequences and again the temperature $\beta=2$ in Eq.~\eqref{eq:max_agg}.
Additional experiment details are in Appendix~\ref{sec:suma_details}.
We test on the standard setting of two news summary datasets: CNN/DailyMail (CNN/DM) \citep{hermann2015teaching} and XSum \citep{narayan2018don}.

\begin{table*}[tb]
\captionsetup{font=footnotesize}
\caption{
\footnotesize
Results on text summarization.
We bold the best result of each metric on each dataset.
The results of Lead-3 on CNN/DM are from \citet{nlpo2022} and on XSum are from \citet{bart2019}. 
Other baseline results are from our reruning RL4LMs' codebase \citep{nlpo2022} using T5-small.
Number reporting formats follow Table~\ref{table:prompt_main}.
} 
\label{table:suma_main}
\centering 
\vspace{-.7em}
 \setlength\tabcolsep{15pt}
\resizebox{\textwidth}{!}{
\footnotesize
\begin{tabular}{@{}lccc|ccc@{}}
\toprule
                  & \multicolumn{3}{c}{CNN/DailyMail}                 & \multicolumn{3}{c}{XSum}                   \\
                  & ROUGE-1      & ROUGE-2      & ROUGE-L      & ROUGE-1      & ROUGE-2      & ROUGE-L      \\ \midrule
Lead-3            &    40.10     &    17.50      &    36.30      & 16.30        & 1.60         & 11.95        \\
Supervised        & 38.88 {\scriptsize (0.02)}    & 16.22 {\scriptsize (0.05)}    & 32.58 {\scriptsize (0.04)}    & 31.79 {\scriptsize (0.02)}    & 9.68 {\scriptsize (0.01)}    & 24.70 {\scriptsize (0.03)}    \\
PPO               & 39.16 {\scriptsize (0.51)}    & 17.37 {\scriptsize (0.33)}    & 33.77 {\scriptsize (0.37)}    & 23.18 {\scriptsize (0.31)}    & 4.46 {\scriptsize (0.19)}    & 16.07 {\scriptsize (0.32)}    \\
Supervised + PPO  & 39.17 {\scriptsize (0.65)} & 17.29 {\scriptsize (0.44)} & 33.76 {\scriptsize (0.53)} & 28.24 {\scriptsize (0.39)}    & 7.68 {\scriptsize (0.13)}    & 20.02 {\scriptsize (0.23)}    \\
NLPO              & 38.90 {\scriptsize (0.35)}    & 17.22 {\scriptsize (0.35)}    & 33.51 {\scriptsize (0.42)}    & 22.97 {\scriptsize (0.23)}    & 4.53 {\scriptsize (0.13)}    & 15.62 {\scriptsize (0.35)}    \\
Supervised + NLPO & 39.27 {\scriptsize (0.60)} & 17.41 {\scriptsize (0.36)} & 33.85 {\scriptsize (0.42)} & 28.08 {\scriptsize (0.16)}    & 7.68 {\scriptsize (0.20)}    & 19.88 {\scriptsize (0.16)}    \\ \midrule
Ours  (AVG)        & \textbf{40.94} {\scriptsize (0.02)} & \textbf{18.78} {\scriptsize (0.03)} & \textbf{38.17} {\scriptsize (0.03)}  & \textbf{33.62} {\scriptsize (0.03)} & \textbf{11.17} {\scriptsize (0.02)} & \textbf{26.33} {\scriptsize (0.05)} \\
Ours  (SUM)        & 40.70 {\scriptsize (0.06)} & 18.48 {\scriptsize (0.05)} & 37.93  {\scriptsize (0.08)}  & 33.27  {\scriptsize (0.09)} & 10.83  {\scriptsize (0.07)} & 25.90  {\scriptsize (0.06)} \\
Ours  (MIN)        &  40.78 {\scriptsize (0.06)} & 18.67  {\scriptsize (0.03)} & 38.01  {\scriptsize (0.04)} & 33.57  {\scriptsize (0.02)} & 11.14  {\scriptsize (0.02)} & 26.30  {\scriptsize (0.03)} \\
Ours  (MAX)        & 39.98  {\scriptsize (0.08)} & 18.06  {\scriptsize (0.03)} & 37.26  {\scriptsize (0.06)} & 32.50  {\scriptsize (0.14)} & 10.46  {\scriptsize (0.12)} & 25.58  {\scriptsize (0.12)} \\ \bottomrule
\end{tabular}
}
\end{table*}

\myparagraph{Results.} \label{sec:main_result_suma}
We compare four variants in our framework with the standard supervised fine-tuning and RL-based methods PPO and NLPO in RL4LMs \citep{nlpo2022} under the environmental reward Meteor --- both with and without task-specific supervised pre-training. 
For a fair comparison, the baseline results are from our rerunning RL4LMs' codebase with a T5-small model as our method.\footnote{We carefully tuned the RL4LMs' baselines on several hyperparameters, which is detailed in Appendix~\ref{sec:suma_details}.}
Table~\ref{table:suma_main} shows the mean and standard deviation of ROUGE-1/2/L score across three random seeds.

On both datasets, our method shows favorable and stable performance against the classical and recent baselines.
The better results of our method over supervised fine-tuning confirm the improvement of
our reward-weighted MLE over the vanilla supervised loss, as discussed in \Secref{sec:policy_training}.
As in the prompt task, the gain of our method over RL-based baselines may indicate the benefit of utilizing our preference-grounded token-level guidance over learning under delayed feedback with KL penalty.  
In this task, using \textit{average} as the aggregation function outperforms the classical \textit{summation}.
This confirms our idea in Section \ref{sec:method_agg_func} on avoiding the interference of unequal sequence-lengths in 
  training $r_\phi$.
Using MIN is also suitable for this task, since it is not confounded by individual lengths and reflects  overall text quality.
Unlike the prompt task, using MAX is unsuitable, since good summaries can hardly be identified by a few keywords.
Overall, these results show the importance of customizing the aggregation choice for the specific LM task, a key feature of our guidance-learning framework.

\begin{figure*}[t]
     \centering
     \begin{subfigure}[b]{0.32\textwidth}
         \centering
         \includegraphics[width=\textwidth]{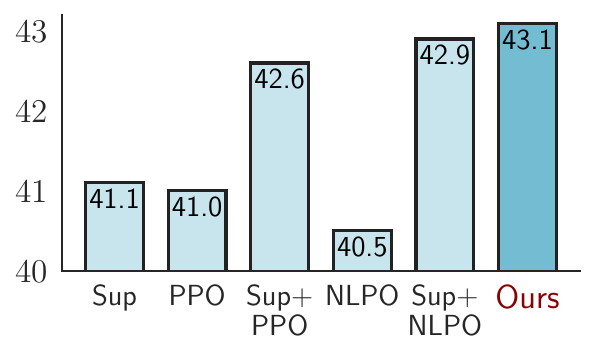}
         \captionsetup{font=footnotesize}
         \vspace{-6mm}
         \caption{\footnotesize{ROUGE-1}}
         \label{fig:t5b_rouge1_bar}
     \end{subfigure}
     \hfill
     \begin{subfigure}[b]{0.32\textwidth}
         \centering
         \includegraphics[width=\textwidth]{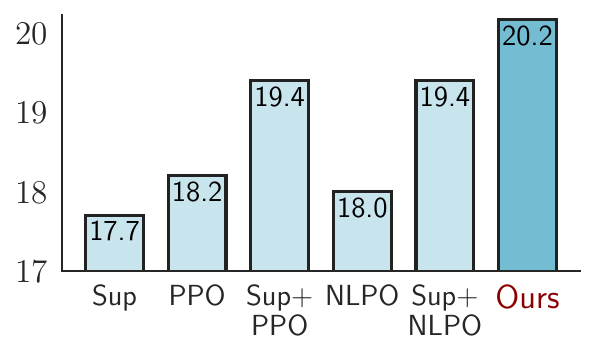}
         \captionsetup{font=footnotesize}
         \vspace{-6mm}
         \caption{\footnotesize{ROUGE-2}}
         \label{fig:t5b_rouge2_bar}
     \end{subfigure}
     \hfill
    \begin{subfigure}[b]{0.32\textwidth}
         \centering
         \includegraphics[width=\textwidth]{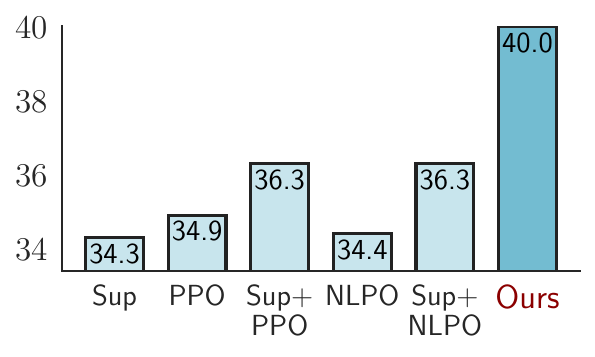}
         \captionsetup{font=footnotesize}
         \vspace{-6mm}
         \caption{\footnotesize{ROUGE-L}}
         \label{fig:t5b_rougel_bar}
     \end{subfigure}
     \\
     \begin{subfigure}[b]{0.32\textwidth}
         \centering
         \includegraphics[width=\textwidth]{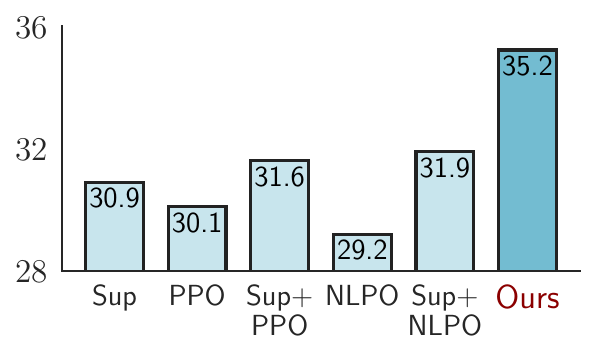}
         \captionsetup{font=footnotesize}
         \vspace{-6mm}
         \caption{\footnotesize{Meteor}}
         \label{fig:t5b_meteor_bar}
         \end{subfigure}
         \hfill
     \begin{subfigure}[b]{0.32\textwidth}
         \centering
         \includegraphics[width=\textwidth]{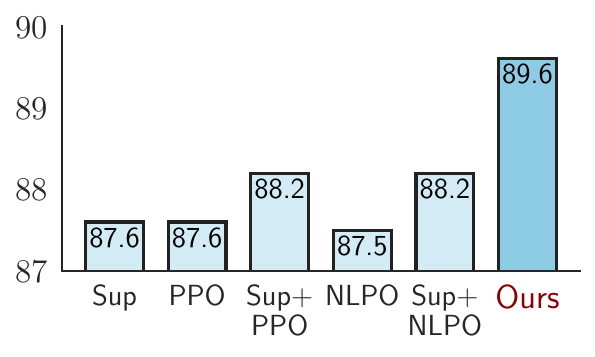}
         \captionsetup{font=footnotesize}
         \vspace{-6mm}
         \caption{\footnotesize{BertScore}}
         \label{fig:t5b_BertScore_bar}
     \end{subfigure}
     \hfill
     \begin{subfigure}[b]{0.32\textwidth}
         \centering
         \includegraphics[width=\textwidth]{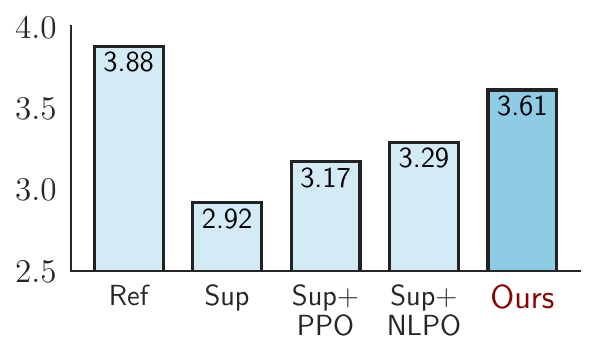}
         \captionsetup{font=footnotesize}
         \vspace{-6mm}
         \caption{\footnotesize{Human}}
         \label{fig:t5b_human_bar}
     \end{subfigure}
     \vspace{-2mm}
     \captionsetup{font=small}
        \caption{ 
        \small
         CNN/DM summarization of our method and baselines under \textbf{T5-base} LM.
        ``Sup'' denotes ``Supervised''.
        ``Ref'' denotes the ground-truth reference summary.
        Except for the human study in \textbf{(f)}, baseline results are directly cited from RL4LMs \citep{nlpo2022} and are the per-metric best across their three environmental rewards. 
        }
        \label{fig:t5b_rouge}
\end{figure*}

Further, to verify the performance of our method under a larger LM, we change the \textit{average} variant of our method in Table~\ref{table:suma_main} from T5-small to using T5-base LM.
Fig.~\ref{fig:t5b_rouge} \textbf{(a)} -- \textbf{(e)} compares our method on CNN/DM against the baselines, with an additional metric BertScore \citep{zhang2019bertscore}. 
The baseline results are directly cited from RL4LMs \citep{nlpo2022} and are the per-metric best across their three environmental rewards.\footnote{The ``ROUGE-L'' here refers to ``Rouge-LSum'' in RL4LMs and HuggingFace, as detailed in Appendix~\ref{sec:suma_details}.}
Table~\ref{table:suma_t5base} in Appendix~\ref{sec:tab_results} shows the detailed numbers.
It is clear that our method performs favorably against these strong baseline methods, especially in the ROUGE-L, BERTScore, Meteor, and ROUGE-2 metrics.
To further varify our method, we conducted a human study under the T5-base LM. 
The results are in Fig.~\ref{fig:t5b_rouge} \textbf{(f)}, with detailed setup and numerics in Table~\ref{table:suma_human} of Appendix~\ref{sec:tab_results}.
It is clear that this human evaluation on the summarization task supports the improvements in ROUGE, Meteor, and BertScore by our method.
Further scaling-up of our method is left as future work.

\subsection{Ablation Study} \label{sec:exp_abla}


This section discusses the following three research questions to better understand our framework. 

\textbf{(a):} \textit{What will be the performance if we switch to using preference-based sequence-level guidance?}

\begin{figure}[t]
     \centering
\includegraphics[width=.88\textwidth]{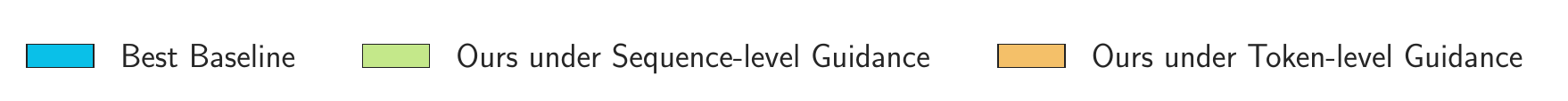}
\\
\vspace{-1.0mm}
     \begin{subfigure}[b]{0.14\textwidth}
         \centering
         \includegraphics[width=\textwidth]{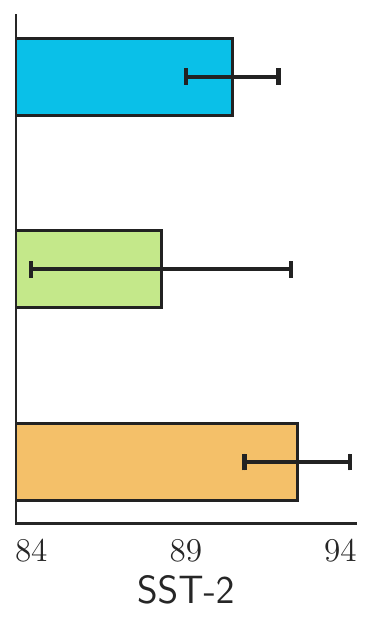}
     \end{subfigure}
     \hfill
     \begin{subfigure}[b]{0.14\textwidth}
         \centering
         \includegraphics[width=\textwidth]{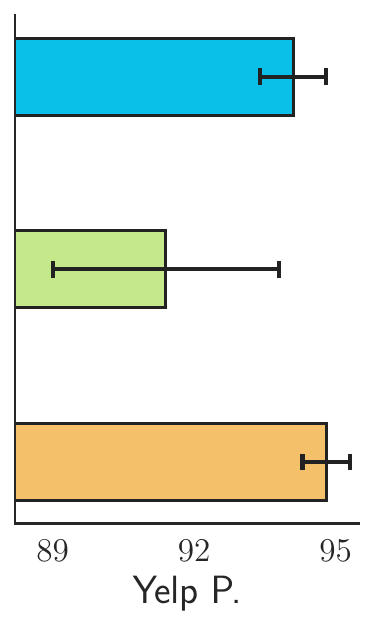}
     \end{subfigure}
     \hfill
    \begin{subfigure}[b]{0.14\textwidth}
         \centering
         \includegraphics[width=\textwidth]{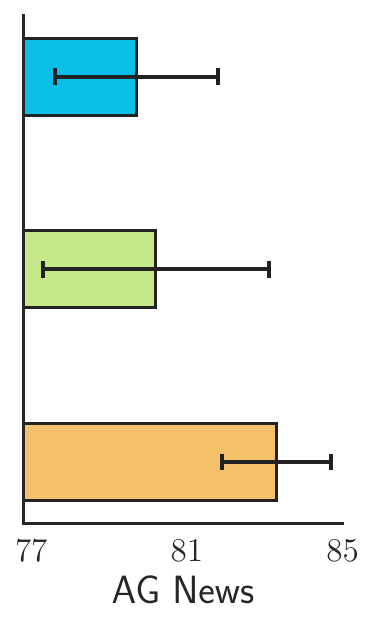}
     \end{subfigure}
     \hfill
    \begin{subfigure}[b]{0.14\textwidth}
         \centering
         \includegraphics[width=\textwidth]{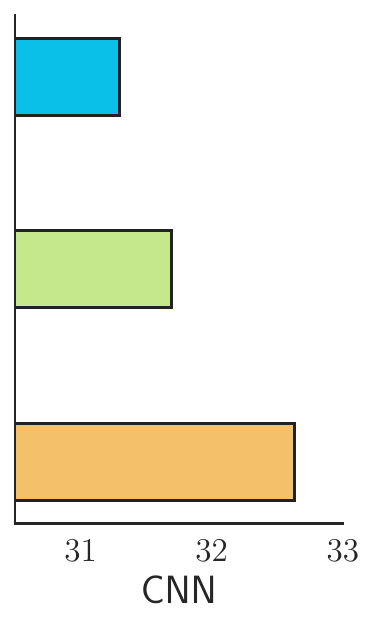}
     \end{subfigure}
     \hfill
    \begin{subfigure}[b]{0.14\textwidth}
         \centering
         \includegraphics[width=\textwidth]{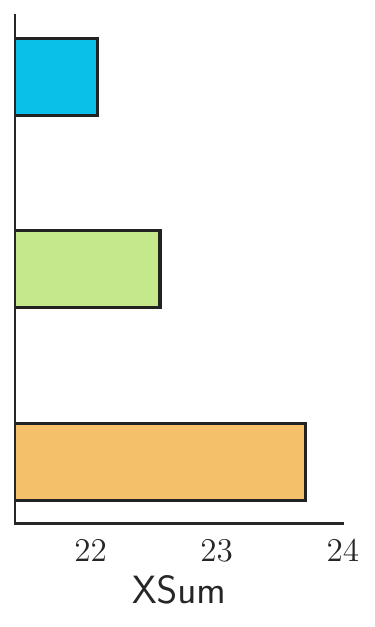}
     \end{subfigure}
     \hfill
    \begin{subfigure}[b]{0.14\textwidth}
         \centering
         \includegraphics[width=\textwidth]{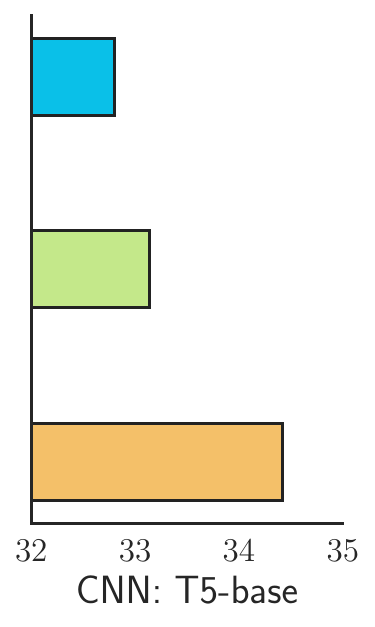}
     \end{subfigure}
     \vspace{-2mm}
     \captionsetup{font=small}
        \caption{ 
        \small
        Performance of our method using sequence-level and token-level preference-based guidance.
        ``Best Baseline'' refers to the best result in the baseline discrete-prompt methods for the prompt task, and the best result over all baseline methods for the summarization task.
        Error bars show one standard deviation.
        }
        \label{fig:seq_rew}
\end{figure}

To further study the gain of grounding preference into token-level guidance,
we change the preference-based \textit{token-level} reward in our method to the corresponding \textit{sequence-level} reward.
Fig.~\ref{fig:seq_rew} shows the results when applying this change to our best variants in the prompt and summarization tasks in Sections \ref{sec:main_result_prompt} and \ref{sec:main_result_suma}, including the T5-base LM in \Secref{sec:main_result_suma}, in comparison to the best corresponding baselines.
For summarization, we plot the average ROUGE scores, \ie, (ROUGE-1 + ROUGE-2 + ROUGE-L) / 3.
Table~\ref{table:seq_rew} in Appendix~\ref{sec:tab_results} shows each  ROUGE metric with standard deviation. 

We see that learning and using preference-based sequence-level guidance does not provide a significant advantage over those baselines that mostly directly work with the task-specific native sequence-level feedback  --- the results are even  much worse than the baselines in some datasets.
Besides, the results of our sequence-level variants are generally less stable. 
These echo the harm of the delayed-feedback issue discussed in \Secref{sec:intro}.
Overall, this set of comparisons confirms that the gain of our framework mainly comes from our preference-grounding  perspective, \ie, learning and using a  preference-based \textit{token-level} guidance, rather than simply learning and using ``a preference-based guidance.''

\textbf{(b):} \textit{How does our method perform if we remove the reward-function retraining scheme?}

\begin{figure}[t]
     \vspace{-1.0mm}
     \centering
\includegraphics[width=.88\textwidth]{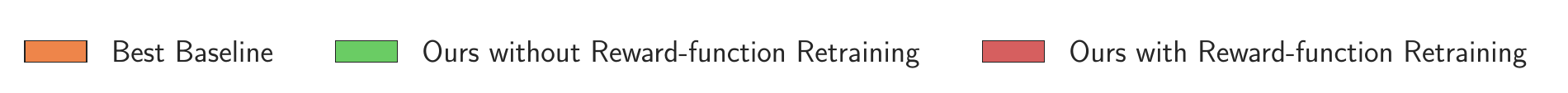}
\\
\vspace{-1.0mm}
     \begin{subfigure}[b]{0.19\textwidth}
         \centering
         \includegraphics[width=\textwidth]{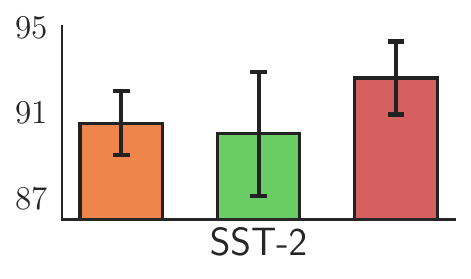}
     \end{subfigure}
     \hfill
     \begin{subfigure}[b]{0.19\textwidth}
         \centering
         \includegraphics[width=\textwidth]{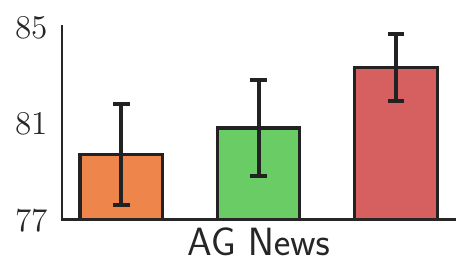}
     \end{subfigure}
     \hfill
    \begin{subfigure}[b]{0.19\textwidth}
         \centering
         \includegraphics[width=\textwidth]{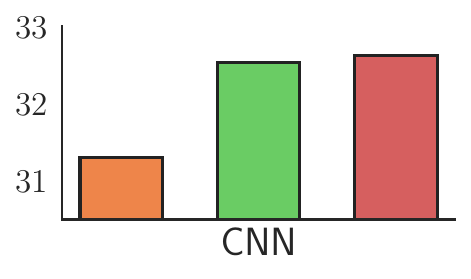}
     \end{subfigure}
     \hfill
    \begin{subfigure}[b]{0.19\textwidth}
         \centering
         \includegraphics[width=\textwidth]{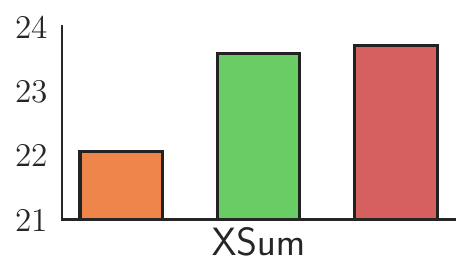}
     \end{subfigure}
     \hfill
    \begin{subfigure}[b]{0.19\textwidth}
         \centering
         \includegraphics[width=\textwidth]{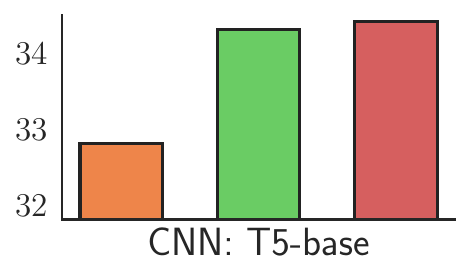}
     \end{subfigure}
     \vspace{-2mm}
     \captionsetup{font=small}
        \caption{ 
        \small
        Performance of our method with and without the reward-function retraining scheme.
        ``Best Baseline'' refers to the same as in Fig.~\ref{fig:seq_rew}.
        Error bars show one standard deviation.
        }
        \label{fig:wo_rew_retrain}
\end{figure}

To study the effect of guidance re-estimation,
we remove the reward-function retraining scheme from our best variants in the prompt and summarization tasks in Sections \ref{sec:main_result_prompt} and \ref{sec:main_result_suma}, including the T5-base LM in \Secref{sec:main_result_suma}.
Fig.~\ref{fig:wo_rew_retrain} compares our methods with the best corresponding baselines.
For the summarization task, we again plot the average ROUGE scores.
Table~\ref{table:wo_rew_retrain} in Appendix~\ref{sec:tab_results} shows each ROUGE metric with standard deviation.
Appendix~\ref{sec:details_rew_retrain_scheme} discusses more  on this re-estimation scheme.

Without guidance re-estimation, our method still performs competitively against the strong baselines, which corroborates the benefit of our preference-grounded guidance.
Fig.~\ref{fig:wo_rew_retrain} also verifies our intuition in \Secref{sec:main_method} that the gain of this scheme depends on the zero-shot ability of the initial LMs.
Specifically, in the prompt task where the initial LM has little zero-shot ability, 
reward-function retraining is helpful to both improve performance and reduce variance. 
In the summarization task where the initial LM does have some zero-shot ability (as shown in \citet{nlpo2022}), guidance re-estimation indeed helps results not as much, since the  distribution-shift issue in \Secref{sec:main_method} is less significant in this case. 
In this task, both our variants, with and without reward retraining, outperform the baselines.

\textbf{(c):} \textit{What if we learn the token-level guidance by a different number of text sequences?}

\begin{figure}[!t]
     \centering
     \begin{subfigure}[b]{0.28\textwidth}
         \centering
         \includegraphics[width=\textwidth]{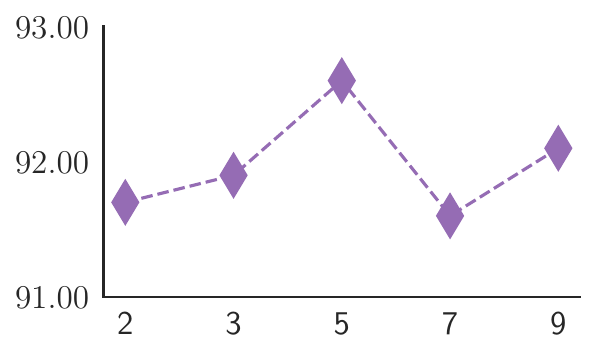}
         \captionsetup{font=footnotesize}
         \vspace{-6mm}
         \caption{\footnotesize{SST-2}} 
         \label{fig:traj_sst_2}
     \end{subfigure}
     \hfill
     \begin{subfigure}[b]{0.28\textwidth}
         \centering
         \includegraphics[width=\textwidth]{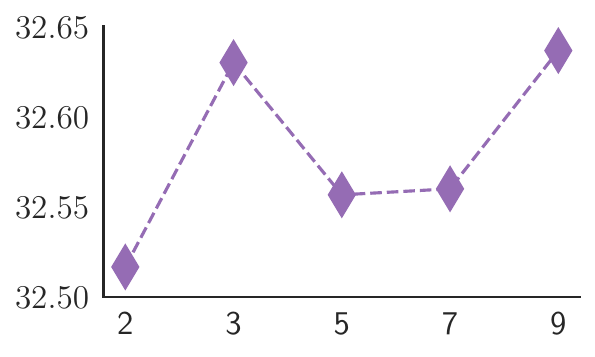}
         \captionsetup{font=footnotesize}
         \vspace{-6mm}
         \caption{\footnotesize{CNN/DM}} 
         \label{fig:traj_cnn}
     \end{subfigure}
     \hfill
    \begin{subfigure}[b]{0.28\textwidth}
         \centering
         \includegraphics[width=\textwidth]{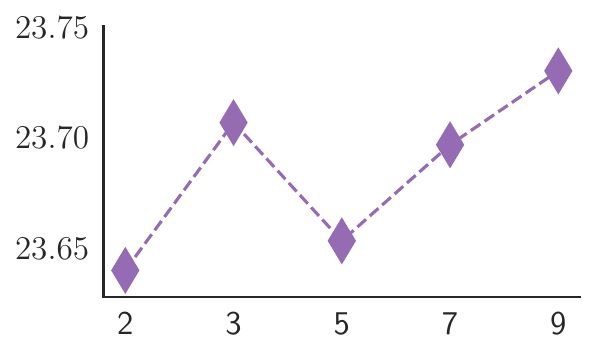}
         \captionsetup{font=footnotesize}
         \vspace{-6mm}
         \caption{\footnotesize{XSum}}    
         \label{fig:traj_xsum}
     \end{subfigure}
     \vspace{-2mm}
     \captionsetup{font=small}
        \caption{ 
        \small
        Varying the number of sequences to learn the token-level guidance, showing mean over random seeds.
        }
        \label{fig:vary_traj}
\end{figure}

To study how the number of sequences used to learn the reward function impacts our method's performance,
we vary this number  in the AVG variant in Tables~\ref{table:prompt_main} and \ref{table:suma_main}.
Fig.~\ref{fig:vary_traj} shows the prompt results on SST-2 and summarization results on CNN/DM and XSum.
For the latter, we again plot the average ROUGE scores.
The scores of each ROUGE metric are in Tables~\ref{table:traj_cnn} and \ref{table:traj_xsum} of Appendix~\ref{sec:tab_results}.

Recall that the best baseline result on SST-2 in Table~\ref{table:prompt_main} is $90.5$, on CNN/DM and XSum in Table~\ref{table:suma_main} is respectively $31.3$ and $22.06$.
Thus, our method is generally robust to the number of sequences used to learn the guidance.
Compared with the classical pairwise-preference learning (\Secref{sec:background}), our framework has the 
flexibility in using multiple sequences.
As illustrated in Fig.~\ref{fig:vary_traj}, 
using three or more sequences to learn the reward function can be generally more beneficial than using only two. 

Due to the page limit, we defer additional ablation study to Appendix~\ref{sec:more_abla}, where  we \textbf{(1)}  show that our framework is generally robust to the hyperparameter $\beta$ in Eq.~\eqref{eq:max_agg} and  $\alpha$ in Eq.~\eqref{eq:prompt_pol_objective}; \textbf{(2)} further validate the harm of the delayed-feedback issue to the relevant LM-training methods on longer text-sequence generation; \textbf{(3)} show that the efficacy of our framework is not tied to the specific preference sources considered in this section.

\section{Conclusion}

To address the granularity mismatch between the sequence-level preference and the token-level LM training losses, in this paper, we develop an alternate-learning process, where we iterate between grounding sequence-level preference into token-level training guidance, and training the LM with the learned guidance.
Our method  performs competitively on two distinct representative LM tasks.
Future work includes combining our preference-grounded guidance with RL-based LM training, and applying our method to human preference and/or other tasks such as (task-oriented) dialog systems.

\begin{ack}
S. Yang, S. Zhang, and M. Zhou acknowledge the support of NSF-IIS 2212418, NIH-R37 CA271186, the Texas Advanced Computing Center (TACC), and the NSF AI Institute for
Foundations of Machine Learning (IFML). S. Yang acknowledges the support of the University Graduate Continuing Fellowship from UT Austin.
\end{ack}
\bibliographystyle{plainnat}
\bibliography{ref}

\begin{thebibliography}{126}
\providecommand{\natexlab}[1]{#1}
\providecommand{\url}[1]{\texttt{#1}}
\expandafter\ifx\csname urlstyle\endcsname\relax
  \providecommand{\doi}[1]{doi: #1}\else
  \providecommand{\doi}{doi: \begingroup \urlstyle{rm}\Url}\fi

\bibitem[An et~al.(2022)An, Li, Lin, Liu, Chen, Fu, Chen, Zheng, and
  Lou]{an2022input}
Shengnan An, Yifei Li, Zeqi Lin, Qian Liu, Bei Chen, Qiang Fu, Weizhu Chen,
  Nanning Zheng, and Jian-Guang Lou.
\newblock Input-tuning: Adapting unfamiliar inputs to frozen pretrained models.
\newblock \emph{arXiv preprint arXiv:2203.03131}, 2022.

\bibitem[Andrychowicz et~al.(2017)Andrychowicz, Wolski, Ray, Schneider, Fong,
  Welinder, McGrew, Tobin, Pieter~Abbeel, and
  Zaremba]{andrychowicz2017hindsight}
Marcin Andrychowicz, Filip Wolski, Alex Ray, Jonas Schneider, Rachel Fong,
  Peter Welinder, Bob McGrew, Josh Tobin, OpenAI Pieter~Abbeel, and Wojciech
  Zaremba.
\newblock Hindsight experience replay.
\newblock \emph{Advances in neural information processing systems}, 30, 2017.

\bibitem[Bai et~al.(2022{\natexlab{a}})Bai, Jones, Ndousse, Askell, Chen,
  DasSarma, Drain, Fort, Ganguli, Henighan, et~al.]{bai2022training}
Yuntao Bai, Andy Jones, Kamal Ndousse, Amanda Askell, Anna Chen, Nova DasSarma,
  Dawn Drain, Stanislav Fort, Deep Ganguli, Tom Henighan, et~al.
\newblock Training a helpful and harmless assistant with reinforcement learning
  from human feedback.
\newblock \emph{arXiv preprint arXiv:2204.05862}, 2022{\natexlab{a}}.

\bibitem[Bai et~al.(2022{\natexlab{b}})Bai, Kadavath, Kundu, Askell, Kernion,
  Jones, Chen, Goldie, Mirhoseini, McKinnon, et~al.]{bai2022constitutional}
Yuntao Bai, Saurav Kadavath, Sandipan Kundu, Amanda Askell, Jackson Kernion,
  Andy Jones, Anna Chen, Anna Goldie, Azalia Mirhoseini, Cameron McKinnon,
  et~al.
\newblock Constitutional ai: Harmlessness from ai feedback.
\newblock \emph{arXiv preprint arXiv:2212.08073}, 2022{\natexlab{b}}.

\bibitem[Banerjee and Lavie(2005)]{meteor2005}
Satanjeev Banerjee and Alon Lavie.
\newblock {METEOR}: An automatic metric for {MT} evaluation with improved
  correlation with human judgments.
\newblock In \emph{Proceedings of the {ACL} Workshop on Intrinsic and Extrinsic
  Evaluation Measures for Machine Translation and/or Summarization}, pages
  65--72, Ann Arbor, Michigan, June 2005. Association for Computational
  Linguistics.
\newblock URL \url{https://aclanthology.org/W05-0909}.

\bibitem[Bengio et~al.(2000)Bengio, Ducharme, and Vincent]{bengio2000neural}
Yoshua Bengio, R{\'e}jean Ducharme, and Pascal Vincent.
\newblock A neural probabilistic language model.
\newblock \emph{Advances in neural information processing systems}, 13, 2000.

\bibitem[Bradley and Terry(1952)]{bradley1952rank}
Ralph~Allan Bradley and Milton~E Terry.
\newblock Rank analysis of incomplete block designs: I. the method of paired
  comparisons.
\newblock \emph{Biometrika}, 39\penalty0 (3/4):\penalty0 324--345, 1952.

\bibitem[Brown et~al.(2019)Brown, Goo, Nagarajan, and Niekum]{trex2019}
Daniel Brown, Wonjoon Goo, Prabhat Nagarajan, and Scott Niekum.
\newblock Extrapolating beyond suboptimal demonstrations via inverse
  reinforcement learning from observations.
\newblock In \emph{International conference on machine learning}, pages
  783--792. PMLR, 2019.

\bibitem[Brown et~al.(2020{\natexlab{a}})Brown, Goo, and Niekum]{drex2020}
Daniel~S Brown, Wonjoon Goo, and Scott Niekum.
\newblock Better-than-demonstrator imitation learning via automatically-ranked
  demonstrations.
\newblock In \emph{Conference on robot learning}, pages 330--359. PMLR,
  2020{\natexlab{a}}.

\bibitem[Brown et~al.(2020{\natexlab{b}})Brown, Mann, Ryder, Subbiah, Kaplan,
  Dhariwal, Neelakantan, Shyam, Sastry, Askell, et~al.]{gpt32020}
Tom Brown, Benjamin Mann, Nick Ryder, Melanie Subbiah, Jared~D Kaplan, Prafulla
  Dhariwal, Arvind Neelakantan, Pranav Shyam, Girish Sastry, Amanda Askell,
  et~al.
\newblock Language models are few-shot learners.
\newblock \emph{Advances in neural information processing systems},
  33:\penalty0 1877--1901, 2020{\natexlab{b}}.

\bibitem[Burges et~al.(2005)Burges, Shaked, Renshaw, Lazier, Deeds, Hamilton,
  and Hullender]{ranknet2005}
Chris Burges, Tal Shaked, Erin Renshaw, Ari Lazier, Matt Deeds, Nicole
  Hamilton, and Greg Hullender.
\newblock Learning to rank using gradient descent.
\newblock In \emph{Proceedings of the 22nd international conference on Machine
  learning}, pages 89--96, 2005.

\bibitem[Castricato et~al.(2022)Castricato, Havrilla, Matiana, Pieler, Ye,
  Yang, Frazier, and Riedl]{castricato2022robust}
Louis Castricato, Alexander Havrilla, Shahbuland Matiana, Michael Pieler,
  Anbang Ye, Ian Yang, Spencer Frazier, and Mark Riedl.
\newblock Robust preference learning for storytelling via contrastive
  reinforcement learning.
\newblock \emph{arXiv preprint arXiv:2210.07792}, 2022.

\bibitem[Christiano et~al.(2017)Christiano, Leike, Brown, Martic, Legg, and
  Amodei]{christiano2017deep}
Paul~F Christiano, Jan Leike, Tom Brown, Miljan Martic, Shane Legg, and Dario
  Amodei.
\newblock Deep reinforcement learning from human preferences.
\newblock \emph{Advances in neural information processing systems}, 30, 2017.

\bibitem[Chung et~al.(2022)Chung, Hou, Longpre, Zoph, Tay, Fedus, Li, Wang,
  Dehghani, Brahma, et~al.]{chung2022scaling}
Hyung~Won Chung, Le~Hou, Shayne Longpre, Barret Zoph, Yi~Tay, William Fedus,
  Eric Li, Xuezhi Wang, Mostafa Dehghani, Siddhartha Brahma, et~al.
\newblock Scaling instruction-finetuned language models.
\newblock \emph{arXiv preprint arXiv:2210.11416}, 2022.

\bibitem[Deng et~al.(2022)Deng, Wang, Hsieh, Wang, Guo, Shu, Song, Xing, and
  Hu]{rlprompt2022}
Mingkai Deng, Jianyu Wang, Cheng-Ping Hsieh, Yihan Wang, Han Guo, Tianmin Shu,
  Meng Song, Eric~P Xing, and Zhiting Hu.
\newblock Rlprompt: Optimizing discrete text prompts with reinforcement
  learning.
\newblock \emph{arXiv preprint arXiv:2205.12548}, 2022.

\bibitem[Devlin et~al.(2018)Devlin, Chang, Lee, and Toutanova]{devlin2018bert}
Jacob Devlin, Ming-Wei Chang, Kenton Lee, and Kristina Toutanova.
\newblock Bert: Pre-training of deep bidirectional transformers for language
  understanding.
\newblock \emph{arXiv preprint arXiv:1810.04805}, 2018.

\bibitem[Diao et~al.(2022)Diao, Li, Lin, Huang, and Zhang]{diao2022black}
Shizhe Diao, Xuechun Li, Yong Lin, Zhichao Huang, and Tong Zhang.
\newblock Black-box prompt learning for pre-trained language models.
\newblock \emph{arXiv preprint arXiv:2201.08531}, 2022.

\bibitem[Durugkar et~al.(2021)Durugkar, Tec, Niekum, and Stone]{aim2021}
Ishan Durugkar, Mauricio Tec, Scott Niekum, and Peter Stone.
\newblock Adversarial intrinsic motivation for reinforcement learning.
\newblock \emph{Advances in Neural Information Processing Systems},
  34:\penalty0 8622--8636, 2021.

\bibitem[Eysenbach et~al.(2021)Eysenbach, Khazatsky, Levine, and
  Salakhutdinov]{mnm2021}
Benjamin Eysenbach, Alexander Khazatsky, Sergey Levine, and Ruslan
  Salakhutdinov.
\newblock {Mismatched No More: Joint Model-Policy Optimization for Model-Based
  RL}.
\newblock \emph{{ArXiv}}, abs/2110.02758, 2021.

\bibitem[Fan et~al.(2020)Fan, Zhang, Chen, and Zhou]{fan2020bayesian}
Xinjie Fan, Shujian Zhang, Bo~Chen, and Mingyuan Zhou.
\newblock Bayesian attention modules.
\newblock \emph{Advances in Neural Information Processing Systems},
  33:\penalty0 16362--16376, 2020.

\bibitem[Feng* et~al.(2023)Feng*, Yang*, Zhang, Zhang, Xiong, Zhou, and
  Wang]{fantasticrewards2022}
Yihao Feng*, Shentao Yang*, Shujian Zhang, Jianguo Zhang, Caiming Xiong,
  Mingyuan Zhou, and Huan Wang.
\newblock Fantastic rewards and how to tame them: A case study on reward
  learning for task-oriented dialogue systems.
\newblock In \emph{The Eleventh International Conference on Learning
  Representations}, 2023.

\bibitem[Freund et~al.(2003)Freund, Iyer, Schapire, and
  Singer]{rankingboosting2003}
Yoav Freund, Raj Iyer, Robert~E Schapire, and Yoram Singer.
\newblock An efficient boosting algorithm for combining preferences.
\newblock \emph{Journal of machine learning research}, 4\penalty0
  (Nov):\penalty0 933--969, 2003.

\bibitem[Fu(2006)]{fu2006gradient}
Michael~C Fu.
\newblock Gradient estimation.
\newblock \emph{Handbooks in operations research and management science},
  13:\penalty0 575--616, 2006.

\bibitem[Gao et~al.(2020)Gao, Fisch, and Chen]{gao2020making}
Tianyu Gao, Adam Fisch, and Danqi Chen.
\newblock Making pre-trained language models better few-shot learners.
\newblock \emph{arXiv preprint arXiv:2012.15723}, 2020.

\bibitem[Ghosh et~al.(2021)Ghosh, Qi, Chaturvedi, and
  Srivastava]{ghosh2021helpful}
Sayan Ghosh, Zheng Qi, Snigdha Chaturvedi, and Shashank Srivastava.
\newblock How helpful is inverse reinforcement learning for table-to-text
  generation?
\newblock In \emph{Proceedings of the 59th Annual Meeting of the Association
  for Computational Linguistics and the 11th International Joint Conference on
  Natural Language Processing (Volume 2: Short Papers)}, pages 71--79, 2021.

\bibitem[Glynn(1990)]{glynn1990likelihood}
Peter~W Glynn.
\newblock Likelihood ratio gradient estimation for stochastic systems.
\newblock \emph{Communications of the ACM}, 33\penalty0 (10):\penalty0 75--84,
  1990.

\bibitem[Go et~al.(2023)Go, Korbak, Kruszewski, Rozen, Ryu, and
  Dymetman]{go2023aligning}
Dongyoung Go, Tomasz Korbak, Germ{\'a}n Kruszewski, Jos Rozen, Nahyeon Ryu, and
  Marc Dymetman.
\newblock Aligning language models with preferences through f-divergence
  minimization.
\newblock \emph{arXiv preprint arXiv:2302.08215}, 2023.

\bibitem[Guo et~al.(2022)Guo, Tan, Liu, Xing, and Hu]{sqltext2021}
Han Guo, Bowen Tan, Zhengzhong Liu, Eric Xing, and Zhiting Hu.
\newblock Efficient (soft) q-learning for text generation with limited good
  data.
\newblock \emph{Findings of the Association for Computational Linguistics:
  EMNLP 2022}, pages 6969--6991, 2022.

\bibitem[Guo et~al.(2018)Guo, Lu, Cai, Zhang, Yu, and Wang]{leakgan2018}
Jiaxian Guo, Sidi Lu, Han Cai, Weinan Zhang, Yong Yu, and Jun Wang.
\newblock Long text generation via adversarial training with leaked
  information.
\newblock In \emph{Proceedings of the AAAI conference on artificial
  intelligence}, volume~32, 2018.

\bibitem[Guo et~al.(2021)Guo, Ainslie, Uthus, Ontanon, Ni, Sung, and
  Yang]{longt52021}
Mandy Guo, Joshua Ainslie, David Uthus, Santiago Ontanon, Jianmo Ni, Yun-Hsuan
  Sung, and Yinfei Yang.
\newblock Longt5: Efficient text-to-text transformer for long sequences.
\newblock \emph{arXiv preprint arXiv:2112.07916}, 2021.

\bibitem[Haarnoja et~al.(2017)Haarnoja, Tang, Abbeel, and
  Levine]{rlenergypolicy2017}
Tuomas Haarnoja, Haoran Tang, P.~Abbeel, and Sergey Levine.
\newblock {Reinforcement Learning with Deep Energy-Based Policies}.
\newblock In \emph{{International Conference on Machine Learning}}, 2017.

\bibitem[Haarnoja et~al.(2018{\natexlab{a}})Haarnoja, Zhou, Abbeel, and
  Levine]{sac2018}
Tuomas Haarnoja, Aurick Zhou, Pieter Abbeel, and Sergey Levine.
\newblock {Soft Actor-Critic: Off-Policy Maximum Entropy Deep Reinforcement
  Learning with a Stochastic {Actor}}.
\newblock In Jennifer Dy and Andreas Krause, editors, \emph{{Proceedings of the
  35th International Conference on Machine Learning}}, volume~80 of
  \emph{Proceedings of Machine Learning Research}, pages 1861--1870. PMLR,
  10--15 Jul 2018{\natexlab{a}}.

\bibitem[Haarnoja et~al.(2018{\natexlab{b}})Haarnoja, Zhou, Hartikainen,
  Tucker, Ha, Tan, Kumar, Zhu, Gupta, Abbeel, and Levine]{sacnew2018}
Tuomas Haarnoja, Aurick Zhou, Kristian Hartikainen, G.~Tucker, Sehoon Ha, Jie
  Tan, Vikash Kumar, Henry Zhu, Abhishek Gupta, P.~Abbeel, and Sergey Levine.
\newblock {Soft Actor-Critic Algorithms and Applications}.
\newblock \emph{ArXiv}, abs/1812.05905, 2018{\natexlab{b}}.

\bibitem[Hambardzumyan et~al.(2021)Hambardzumyan, Khachatrian, and
  May]{hambardzumyan2021warp}
Karen Hambardzumyan, Hrant Khachatrian, and Jonathan May.
\newblock Warp: Word-level adversarial reprogramming.
\newblock \emph{arXiv preprint arXiv:2101.00121}, 2021.

\bibitem[Hancock et~al.(2019)Hancock, Bordes, Mazare, and
  Weston]{hancock2019learning}
Braden Hancock, Antoine Bordes, Pierre-Emmanuel Mazare, and Jason Weston.
\newblock Learning from dialogue after deployment: Feed yourself, chatbot!
\newblock \emph{arXiv preprint arXiv:1901.05415}, 2019.

\bibitem[Hasan et~al.(2021)Hasan, Bhattacharjee, Islam, Mubasshir, Li, Kang,
  Rahman, and Shahriyar]{mt52021}
Tahmid Hasan, Abhik Bhattacharjee, Md.~Saiful Islam, Kazi Mubasshir, Yuan-Fang
  Li, Yong-Bin Kang, M.~Sohel Rahman, and Rifat Shahriyar.
\newblock {XL}-sum: Large-scale multilingual abstractive summarization for 44
  languages.
\newblock In \emph{Findings of the Association for Computational Linguistics:
  ACL-IJCNLP 2021}, pages 4693--4703, Online, August 2021. Association for
  Computational Linguistics.
\newblock \doi{10.18653/v1/2021.findings-acl.413}.
\newblock URL \url{https://aclanthology.org/2021.findings-acl.413}.

\bibitem[Herbrich et~al.(1999)Herbrich, Graepel, and Obermayer]{ranksvm1999}
Ralf Herbrich, Thore Graepel, and Klaus Obermayer.
\newblock Support vector learning for ordinal regression.
\newblock \emph{IET}, 1999.

\bibitem[Hermann et~al.(2015)Hermann, Kocisky, Grefenstette, Espeholt, Kay,
  Suleyman, and Blunsom]{hermann2015teaching}
Karl~Moritz Hermann, Tomas Kocisky, Edward Grefenstette, Lasse Espeholt, Will
  Kay, Mustafa Suleyman, and Phil Blunsom.
\newblock Teaching machines to read and comprehend.
\newblock \emph{Advances in neural information processing systems}, 28, 2015.

\bibitem[Hishinuma and Senda(2021)]{wmopo2021}
Toru Hishinuma and Kei Senda.
\newblock {Weighted model estimation for offline model-based reinforcement
  learning}.
\newblock In \emph{{Advances in neural information processing systems}}, 2021.

\bibitem[Jaques et~al.(2019)Jaques, Ghandeharioun, Shen, Ferguson, Lapedriza,
  Jones, Gu, and Picard]{offlinerldialog2019}
Natasha Jaques, Asma Ghandeharioun, Judy~Hanwen Shen, Craig Ferguson,
  {\`A}.~Lapedriza, Noah~J. Jones, S.~Gu, and Rosalind~W. Picard.
\newblock {Way Off-Policy Batch Deep Reinforcement Learning of Implicit Human
  Preferences in Dialog}.
\newblock \emph{ArXiv}, abs/1907.00456, 2019.

\bibitem[Jaques et~al.(2020)Jaques, Shen, Ghandeharioun, Ferguson, Lapedriza,
  Jones, Gu, and Picard]{jaques2020human}
Natasha Jaques, Judy~Hanwen Shen, Asma Ghandeharioun, Craig Ferguson, Agata
  Lapedriza, Noah Jones, Shixiang~Shane Gu, and Rosalind Picard.
\newblock Human-centric dialog training via offline reinforcement learning.
\newblock \emph{arXiv preprint arXiv:2010.05848}, 2020.

\bibitem[Junczys-Dowmunt et~al.(2018)Junczys-Dowmunt, Grundkiewicz, Guha, and
  Heafield]{Junczys2018approaching}
Marcin Junczys-Dowmunt, Roman Grundkiewicz, Shubha Guha, and Kenneth Heafield.
\newblock Approaching neural grammatical error correction as a low-resource
  machine translation task.
\newblock \emph{arXiv preprint arXiv:1804.05940}, 2018.

\bibitem[Khalifa et~al.(2021)Khalifa, Elsahar, and Dymetman]{khalifa2021a}
Muhammad Khalifa, Hady Elsahar, and Marc Dymetman.
\newblock A distributional approach to controlled text generation.
\newblock In \emph{International Conference on Learning Representations}, 2021.
\newblock URL \url{https://openreview.net/forum?id=jWkw45-9AbL}.

\bibitem[Khashabi et~al.(2021)Khashabi, Lyu, Min, Qin, Richardson, Singh,
  Welleck, Hajishirzi, Khot, Sabharwal, et~al.]{khashabi2021prompt}
Daniel Khashabi, Shane Lyu, Sewon Min, Lianhui Qin, Kyle Richardson, Sameer
  Singh, Sean Welleck, Hannaneh Hajishirzi, Tushar Khot, Ashish Sabharwal,
  et~al.
\newblock Prompt waywardness: The curious case of discretized interpretation of
  continuous prompts.
\newblock \emph{arXiv preprint arXiv:2112.08348}, 2021.

\bibitem[Kim et~al.(2023)Kim, Park, Shin, Lee, Abbeel, and
  Lee]{kim2023preference}
Changyeon Kim, Jongjin Park, Jinwoo Shin, Honglak Lee, Pieter Abbeel, and Kimin
  Lee.
\newblock Preference transformer: Modeling human preferences using transformers
  for {RL}.
\newblock In \emph{The Eleventh International Conference on Learning
  Representations}, 2023.
\newblock URL \url{https://openreview.net/forum?id=Peot1SFDX0}.

\bibitem[Kingma and Ba(2014)]{adam2014}
Diederik~P. Kingma and Jimmy Ba.
\newblock {Adam: A Method for Stochastic Optimization}.
\newblock In \emph{{International Conference on Learning Representations}},
  2014.

\bibitem[Korbak et~al.(2022)Korbak, Elsahar, Kruszewski, and
  Dymetman]{korbak2022reinforcement}
Tomasz Korbak, Hady Elsahar, Germ{\'a}n Kruszewski, and Marc Dymetman.
\newblock On reinforcement learning and distribution matching for fine-tuning
  language models with no catastrophic forgetting.
\newblock \emph{arXiv preprint arXiv:2206.00761}, 2022.

\bibitem[Korbak et~al.(2023)Korbak, Shi, Chen, Bhalerao, Buckley, Phang,
  Bowman, and Perez]{korbak2023pretraining}
Tomasz Korbak, Kejian Shi, Angelica Chen, Rasika Bhalerao, Christopher~L
  Buckley, Jason Phang, Samuel~R Bowman, and Ethan Perez.
\newblock Pretraining language models with human preferences.
\newblock \emph{arXiv preprint arXiv:2302.08582}, 2023.

\bibitem[Le et~al.(2022)Le, Wang, Gotmare, Savarese, and Hoi]{le2022coderl}
Hung Le, Yue Wang, Akhilesh~Deepak Gotmare, Silvio Savarese, and Steven
  Chu~Hong Hoi.
\newblock Coderl: Mastering code generation through pretrained models and deep
  reinforcement learning.
\newblock \emph{Advances in Neural Information Processing Systems},
  35:\penalty0 21314--21328, 2022.

\bibitem[Lester et~al.(2021)Lester, Al-Rfou, and Constant]{lester2021power}
Brian Lester, Rami Al-Rfou, and Noah Constant.
\newblock The power of scale for parameter-efficient prompt tuning.
\newblock \emph{arXiv preprint arXiv:2104.08691}, 2021.

\bibitem[Lewis et~al.(2019)Lewis, Liu, Goyal, Ghazvininejad, Mohamed, Levy,
  Stoyanov, and Zettlemoyer]{bart2019}
Mike Lewis, Yinhan Liu, Naman Goyal, Marjan Ghazvininejad, Abdelrahman Mohamed,
  Omer Levy, Ves Stoyanov, and Luke Zettlemoyer.
\newblock Bart: Denoising sequence-to-sequence pre-training for natural
  language generation, translation, and comprehension.
\newblock \emph{arXiv preprint arXiv:1910.13461}, 2019.

\bibitem[Li and Liang(2021)]{prefixtuning2021}
Xiang~Lisa Li and Percy Liang.
\newblock Prefix-tuning: Optimizing continuous prompts for generation.
\newblock \emph{arXiv preprint arXiv:2101.00190}, 2021.

\bibitem[Lin(2004)]{rouge2004}
Chin-Yew Lin.
\newblock {ROUGE}: A package for automatic evaluation of summaries.
\newblock In \emph{Text Summarization Branches Out}, pages 74--81, Barcelona,
  Spain, July 2004. Association for Computational Linguistics.
\newblock URL \url{https://aclanthology.org/W04-1013}.

\bibitem[Lin et~al.(2017)Lin, Li, He, Zhang, and Sun]{adversarialranking2017}
Kevin Lin, Dianqi Li, Xiaodong He, Zhengyou Zhang, and Ming-Ting Sun.
\newblock Adversarial ranking for language generation.
\newblock \emph{Advances in neural information processing systems}, 30, 2017.

\bibitem[Liu et~al.(2019{\natexlab{a}})Liu, Trott, Socher, and
  Xiong]{liu2018competitive}
Hao Liu, Alexander Trott, Richard Socher, and Caiming Xiong.
\newblock Competitive experience replay.
\newblock In \emph{International Conference on Learning Representations},
  2019{\natexlab{a}}.

\bibitem[Liu et~al.(2021{\natexlab{a}})Liu, Yuan, Fu, Jiang, Hayashi, and
  Neubig]{promptsurvey2021}
Pengfei Liu, Weizhe Yuan, Jinlan Fu, Zhengbao Jiang, Hiroaki Hayashi, and
  Graham Neubig.
\newblock Pre-train, prompt, and predict: A systematic survey of prompting
  methods in natural language processing.
\newblock \emph{arXiv preprint arXiv:2107.13586}, 2021{\natexlab{a}}.

\bibitem[Liu et~al.(2021{\natexlab{b}})Liu, Zheng, Du, Ding, Qian, Yang, and
  Tang]{liu2021gpt}
Xiao Liu, Yanan Zheng, Zhengxiao Du, Ming Ding, Yujie Qian, Zhilin Yang, and
  Jie Tang.
\newblock Gpt understands, too.
\newblock \emph{arXiv preprint arXiv:2103.10385}, 2021{\natexlab{b}}.

\bibitem[Liu et~al.(2019{\natexlab{b}})Liu, Ott, Goyal, Du, Joshi, Chen, Levy,
  Lewis, Zettlemoyer, and Stoyanov]{liu2019roberta}
Yinhan Liu, Myle Ott, Naman Goyal, Jingfei Du, Mandar Joshi, Danqi Chen, Omer
  Levy, Mike Lewis, Luke Zettlemoyer, and Veselin Stoyanov.
\newblock Roberta: A robustly optimized bert pretraining approach.
\newblock \emph{arXiv preprint arXiv:1907.11692}, 2019{\natexlab{b}}.

\bibitem[Loshchilov and Hutter(2017)]{adamw2017}
Ilya Loshchilov and Frank Hutter.
\newblock Decoupled weight decay regularization.
\newblock \emph{arXiv preprint arXiv:1711.05101}, 2017.

\bibitem[Lu et~al.(2022)Lu, Welleck, Jiang, Hessel, Qin, West, Ammanabrolu, and
  Choi]{quark2022}
Ximing Lu, Sean Welleck, Liwei Jiang, Jack Hessel, Lianhui Qin, Peter West,
  Prithviraj Ammanabrolu, and Yejin Choi.
\newblock Quark: Controllable text generation with reinforced unlearning.
\newblock \emph{arXiv preprint arXiv:2205.13636}, 2022.

\bibitem[Luce(2012)]{luce2012individual}
R~Duncan Luce.
\newblock \emph{Individual choice behavior: A theoretical analysis}.
\newblock Courier Corporation, 2012.

\bibitem[Menick et~al.(2022)Menick, Trebacz, Mikulik, Aslanides, Song,
  Chadwick, Glaese, Young, Campbell-Gillingham, Irving,
  et~al.]{menick2022teaching}
Jacob Menick, Maja Trebacz, Vladimir Mikulik, John Aslanides, Francis Song,
  Martin Chadwick, Mia Glaese, Susannah Young, Lucy Campbell-Gillingham,
  Geoffrey Irving, et~al.
\newblock Teaching language models to support answers with verified quotes.
\newblock \emph{arXiv preprint arXiv:2203.11147}, 2022.

\bibitem[Mishra et~al.(2021)Mishra, Khashabi, Baral, and
  Hajishirzi]{mishra2021cross}
Swaroop Mishra, Daniel Khashabi, Chitta Baral, and Hannaneh Hajishirzi.
\newblock Cross-task generalization via natural language crowdsourcing
  instructions.
\newblock \emph{arXiv preprint arXiv:2104.08773}, 2021.

\bibitem[Narayan et~al.(2018)Narayan, Cohen, and Lapata]{narayan2018don}
Shashi Narayan, Shay~B Cohen, and Mirella Lapata.
\newblock Don't give me the details, just the summary! topic-aware
  convolutional neural networks for extreme summarization.
\newblock \emph{arXiv preprint arXiv:1808.08745}, 2018.

\bibitem[Norouzi et~al.(2016)Norouzi, Bengio, Jaitly, Schuster, Wu, Schuurmans,
  et~al.]{norouzi2016reward}
Mohammad Norouzi, Samy Bengio, Navdeep Jaitly, Mike Schuster, Yonghui Wu, Dale
  Schuurmans, et~al.
\newblock Reward augmented maximum likelihood for neural structured prediction.
\newblock \emph{Advances In Neural Information Processing Systems}, 29, 2016.

\bibitem[OpenAI(2023)]{gpt42023}
OpenAI.
\newblock Gpt-4 technical report, 2023.

\bibitem[Ouyang et~al.(2022)Ouyang, Wu, Jiang, Almeida, Wainwright, Mishkin,
  Zhang, Agarwal, Slama, Ray, et~al.]{instructgpt2022}
Long Ouyang, Jeff Wu, Xu~Jiang, Diogo Almeida, Carroll~L Wainwright, Pamela
  Mishkin, Chong Zhang, Sandhini Agarwal, Katarina Slama, Alex Ray, et~al.
\newblock Training language models to follow instructions with human feedback.
\newblock \emph{arXiv preprint arXiv:2203.02155}, 2022.

\bibitem[Pang and He(2020)]{pang2020text}
Richard~Yuanzhe Pang and He~He.
\newblock Text generation by learning from demonstrations.
\newblock \emph{arXiv preprint arXiv:2009.07839}, 2020.

\bibitem[Pang et~al.(2022)Pang, Padmakumar, Sellam, Parikh, and
  He]{rewardgaming2022}
Yuanzhe~Richard Pang, Vishakh Padmakumar, Thibault Sellam, Ankur~P Parikh, and
  He~He.
\newblock Reward gaming in conditional text generation.
\newblock \emph{arXiv e-prints}, pages arXiv--2211, 2022.

\bibitem[Paulus et~al.(2017)Paulus, Xiong, and Socher]{paulus2017deep}
Romain Paulus, Caiming Xiong, and Richard Socher.
\newblock A deep reinforced model for abstractive summarization.
\newblock \emph{arXiv preprint arXiv:1705.04304}, 2017.

\bibitem[Perez et~al.(2021)Perez, Kiela, and Cho]{perez2021true}
Ethan Perez, Douwe Kiela, and Kyunghyun Cho.
\newblock True few-shot learning with language models.
\newblock \emph{Advances in Neural Information Processing Systems},
  34:\penalty0 11054--11070, 2021.

\bibitem[Plackett(1975)]{plackett1975analysis}
Robin~L Plackett.
\newblock The analysis of permutations.
\newblock \emph{Journal of the Royal Statistical Society: Series C (Applied
  Statistics)}, 24\penalty0 (2):\penalty0 193--202, 1975.

\bibitem[Prasad et~al.(2022)Prasad, Hase, Zhou, and Bansal]{prasad2022grips}
Archiki Prasad, Peter Hase, Xiang Zhou, and Mohit Bansal.
\newblock Grips: Gradient-free, edit-based instruction search for prompting
  large language models.
\newblock \emph{arXiv preprint arXiv:2203.07281}, 2022.

\bibitem[Qin and Eisner(2021)]{qin2021learning}
Guanghui Qin and Jason Eisner.
\newblock Learning how to ask: Querying lms with mixtures of soft prompts.
\newblock \emph{arXiv preprint arXiv:2104.06599}, 2021.

\bibitem[Radford and Sutskever(2018)]{gpt2018}
Alec Radford and Ilya Sutskever.
\newblock {Improving Language Understanding by Generative Pre-Training}.
\newblock In \emph{{arxiv}}, 2018.

\bibitem[Radford et~al.(2019)Radford, Wu, Child, Luan, Amodei, Sutskever,
  et~al.]{gpt2}
Alec Radford, Jeffrey Wu, Rewon Child, David Luan, Dario Amodei, Ilya
  Sutskever, et~al.
\newblock Language models are unsupervised multitask learners.
\newblock \emph{OpenAI blog}, 1\penalty0 (8):\penalty0 9, 2019.

\bibitem[Raffel et~al.(2020)Raffel, Shazeer, Roberts, Lee, Narang, Matena,
  Zhou, Li, Liu, et~al.]{t52020}
Colin Raffel, Noam Shazeer, Adam Roberts, Katherine Lee, Sharan Narang, Michael
  Matena, Yanqi Zhou, Wei Li, Peter~J Liu, et~al.
\newblock Exploring the limits of transfer learning with a unified text-to-text
  transformer.
\newblock \emph{J. Mach. Learn. Res.}, 21\penalty0 (140):\penalty0 1--67, 2020.

\bibitem[Ramachandran et~al.(2021)Ramachandran, Hashimoto, and
  Xiong]{caspi2021}
Govardana~Sachithanandam Ramachandran, Kazuma Hashimoto, and Caiming Xiong.
\newblock Causal-aware safe policy improvement for task-oriented dialogue.
\newblock \emph{arXiv preprint arXiv:2103.06370}, 2021.

\bibitem[Ramamurthy et~al.(2022)Ramamurthy, Ammanabrolu, Brantley, Hessel,
  Sifa, Bauckhage, Hajishirzi, and Choi]{nlpo2022}
Rajkumar Ramamurthy, Prithviraj Ammanabrolu, Kiant{\'e} Brantley, Jack Hessel,
  Rafet Sifa, Christian Bauckhage, Hannaneh Hajishirzi, and Yejin Choi.
\newblock Is reinforcement learning (not) for natural language processing?:
  Benchmarks, baselines, and building blocks for natural language policy
  optimization.
\newblock \emph{arXiv preprint arXiv:2210.01241}, 2022.

\bibitem[Ranzato et~al.(2015)Ranzato, Chopra, Auli, and
  Zaremba]{ranzato2015sequence}
Marc'Aurelio Ranzato, Sumit Chopra, Michael Auli, and Wojciech Zaremba.
\newblock Sequence level training with recurrent neural networks.
\newblock \emph{arXiv preprint arXiv:1511.06732}, 2015.

\bibitem[Rennie et~al.(2017)Rennie, Marcheret, Mroueh, Ross, and
  Goel]{rennie2017self}
Steven~J Rennie, Etienne Marcheret, Youssef Mroueh, Jerret Ross, and Vaibhava
  Goel.
\newblock Self-critical sequence training for image captioning.
\newblock In \emph{Proceedings of the IEEE conference on computer vision and
  pattern recognition}, pages 7008--7024, 2017.

\bibitem[Ryang and Abekawa(2012)]{ryang2012framework}
Seonggi Ryang and Takeshi Abekawa.
\newblock Framework of automatic text summarization using reinforcement
  learning.
\newblock In \emph{Proceedings of the 2012 Joint Conference on Empirical
  Methods in Natural Language Processing and Computational Natural Language
  Learning}, pages 256--265, Jeju Island, Korea, July 2012. Association for
  Computational Linguistics.
\newblock URL \url{https://aclanthology.org/D12-1024}.

\bibitem[Sanh et~al.(2021)Sanh, Webson, Raffel, Bach, Sutawika, Alyafeai,
  Chaffin, Stiegler, Scao, Raja, et~al.]{sanh2021multitask}
Victor Sanh, Albert Webson, Colin Raffel, Stephen~H Bach, Lintang Sutawika,
  Zaid Alyafeai, Antoine Chaffin, Arnaud Stiegler, Teven~Le Scao, Arun Raja,
  et~al.
\newblock Multitask prompted training enables zero-shot task generalization.
\newblock \emph{arXiv preprint arXiv:2110.08207}, 2021.

\bibitem[Sanh et~al.(2022)Sanh, Webson, Raffel, Bach, Sutawika, Alyafeai,
  Chaffin, Stiegler, Raja, Dey, Bari, Xu, Thakker, Sharma, Szczechla, Kim,
  Chhablani, Nayak, Datta, Chang, Jiang, Wang, Manica, Shen, Yong, Pandey,
  Bawden, Wang, Neeraj, Rozen, Sharma, Santilli, Fevry, Fries, Teehan, Scao,
  Biderman, Gao, Wolf, and Rush]{sanh2022multitask}
Victor Sanh, Albert Webson, Colin Raffel, Stephen Bach, Lintang Sutawika, Zaid
  Alyafeai, Antoine Chaffin, Arnaud Stiegler, Arun Raja, Manan Dey, M~Saiful
  Bari, Canwen Xu, Urmish Thakker, Shanya~Sharma Sharma, Eliza Szczechla,
  Taewoon Kim, Gunjan Chhablani, Nihal Nayak, Debajyoti Datta, Jonathan Chang,
  Mike Tian-Jian Jiang, Han Wang, Matteo Manica, Sheng Shen, Zheng~Xin Yong,
  Harshit Pandey, Rachel Bawden, Thomas Wang, Trishala Neeraj, Jos Rozen,
  Abheesht Sharma, Andrea Santilli, Thibault Fevry, Jason~Alan Fries, Ryan
  Teehan, Teven~Le Scao, Stella Biderman, Leo Gao, Thomas Wolf, and Alexander~M
  Rush.
\newblock Multitask prompted training enables zero-shot task generalization.
\newblock In \emph{International Conference on Learning Representations}, 2022.
\newblock URL \url{https://openreview.net/forum?id=9Vrb9D0WI4}.

\bibitem[Scheurer et~al.(2022)Scheurer, Campos, Chan, Chen, Cho, and
  Perez]{scheurer2022training}
Jérémy Scheurer, Jon~Ander Campos, Jun~Shern Chan, Angelica Chen, Kyunghyun
  Cho, and Ethan Perez.
\newblock Training language models with language feedback, 2022.

\bibitem[Schick and Sch{\"u}tze(2020{\natexlab{a}})]{schick2020exploiting}
Timo Schick and Hinrich Sch{\"u}tze.
\newblock Exploiting cloze questions for few shot text classification and
  natural language inference.
\newblock \emph{arXiv preprint arXiv:2001.07676}, 2020{\natexlab{a}}.

\bibitem[Schick and Sch{\"u}tze(2020{\natexlab{b}})]{schick2020s}
Timo Schick and Hinrich Sch{\"u}tze.
\newblock It's not just size that matters: Small language models are also
  few-shot learners.
\newblock \emph{arXiv preprint arXiv:2009.07118}, 2020{\natexlab{b}}.

\bibitem[Shi et~al.(2018)Shi, Chen, Qiu, and Huang]{shi2018toward}
Zhan Shi, Xinchi Chen, Xipeng Qiu, and Xuanjing Huang.
\newblock Toward diverse text generation with inverse reinforcement learning.
\newblock \emph{arXiv preprint arXiv:1804.11258}, 2018.

\bibitem[Shin et~al.(2020)Shin, Razeghi, Logan~IV, Wallace, and
  Singh]{shin2020autoprompt}
Taylor Shin, Yasaman Razeghi, Robert~L Logan~IV, Eric Wallace, and Sameer
  Singh.
\newblock Autoprompt: Eliciting knowledge from language models with
  automatically generated prompts.
\newblock \emph{arXiv preprint arXiv:2010.15980}, 2020.

\bibitem[Shu et~al.(2021)Shu, Yoo, and Ha]{shu2021reward}
Raphael Shu, Kang~Min Yoo, and Jung-Woo Ha.
\newblock Reward optimization for neural machine translation with learned
  metrics.
\newblock \emph{arXiv preprint arXiv:2104.07541}, 2021.

\bibitem[Snell et~al.(2022)Snell, Kostrikov, Su, Yang, and
  Levine]{snell2022offline}
Charlie Snell, Ilya Kostrikov, Yi~Su, Mengjiao Yang, and Sergey Levine.
\newblock Offline rl for natural language generation with implicit language q
  learning.
\newblock \emph{arXiv preprint arXiv:2206.11871}, 2022.

\bibitem[Socher et~al.(2013)Socher, Perelygin, Wu, Chuang, Manning, Ng, and
  Potts]{socher2013recursive}
Richard Socher, Alex Perelygin, Jean Wu, Jason Chuang, Christopher~D Manning,
  Andrew~Y Ng, and Christopher Potts.
\newblock Recursive deep models for semantic compositionality over a sentiment
  treebank.
\newblock In \emph{Proceedings of the 2013 conference on empirical methods in
  natural language processing}, pages 1631--1642, 2013.

\bibitem[Solaiman and Dennison(2021)]{solaiman2021process}
Irene Solaiman and Christy Dennison.
\newblock Process for adapting language models to society ({PALMS}) with
  values-targeted datasets.
\newblock In A.~Beygelzimer, Y.~Dauphin, P.~Liang, and J.~Wortman Vaughan,
  editors, \emph{Advances in Neural Information Processing Systems}, 2021.
\newblock URL \url{https://openreview.net/forum?id=k-ghaB9VZBw}.

\bibitem[Stiennon et~al.(2020)Stiennon, Ouyang, Wu, Ziegler, Lowe, Voss,
  Radford, Amodei, and Christiano]{stiennon2020learning}
Nisan Stiennon, Long Ouyang, Jeffrey Wu, Daniel Ziegler, Ryan Lowe, Chelsea
  Voss, Alec Radford, Dario Amodei, and Paul~F Christiano.
\newblock Learning to summarize from human feedback.
\newblock \emph{Advances in Neural Information Processing Systems},
  33:\penalty0 3008--3021, 2020.

\bibitem[Su et~al.(2021)Su, Wang, Qin, Chan, Lin, Liu, Li, Li, Hou, Sun,
  et~al.]{su2021transferability}
Yusheng Su, Xiaozhi Wang, Yujia Qin, Chi-Min Chan, Yankai Lin, Zhiyuan Liu,
  Peng Li, Juanzi Li, Lei Hou, Maosong Sun, et~al.
\newblock On transferability of prompt tuning for natural language
  understanding.
\newblock \emph{arXiv preprint arXiv:2111.06719}, 2021.

\bibitem[Sun et~al.(2022)Sun, Shao, Qian, Huang, and Qiu]{sun2022black}
Tianxiang Sun, Yunfan Shao, Hong Qian, Xuanjing Huang, and Xipeng Qiu.
\newblock Black-box tuning for language-model-as-a-service.
\newblock \emph{arXiv preprint arXiv:2201.03514}, 2022.

\bibitem[Sutton and Barto(2018)]{rlintro2018}
Richard~S Sutton and Andrew~G Barto.
\newblock \emph{Reinforcement learning: An introduction}.
\newblock MIT press, 2018.

\bibitem[Takanobu et~al.(2019)Takanobu, Zhu, and Huang]{takanobu2019guided}
Ryuichi Takanobu, Hanlin Zhu, and Minlie Huang.
\newblock Guided dialog policy learning: Reward estimation for multi-domain
  task-oriented dialog.
\newblock \emph{arXiv preprint arXiv:1908.10719}, 2019.

\bibitem[Todorov et~al.(2012)Todorov, Erez, and Tassa]{todorov2012mujoco}
Emanuel Todorov, Tom Erez, and Yuval Tassa.
\newblock Mujoco: A physics engine for model-based control.
\newblock In \emph{2012 IEEE/RSJ International Conference on Intelligent Robots
  and Systems}, pages 5026--5033. IEEE, 2012.
\newblock \doi{10.1109/IROS.2012.6386109}.

\bibitem[Vaswani et~al.(2017)Vaswani, Shazeer, Parmar, Uszkoreit, Jones, Gomez,
  Kaiser, and Polosukhin]{vaswani2017attention}
Ashish Vaswani, Noam Shazeer, Niki Parmar, Jakob Uszkoreit, Llion Jones,
  Aidan~N Gomez, {\L}ukasz Kaiser, and Illia Polosukhin.
\newblock Attention is all you need.
\newblock \emph{Advances in neural information processing systems}, 30, 2017.

\bibitem[Wang et~al.(2020{\natexlab{a}})Wang, Peng, and Wong]{wang2020learning}
Huimin Wang, Baolin Peng, and Kam-Fai Wong.
\newblock Learning efficient dialogue policy from demonstrations through
  shaping.
\newblock In \emph{Proceedings of the 58th Annual Meeting of the Association
  for Computational Linguistics}, pages 6355--6365, 2020{\natexlab{a}}.

\bibitem[Wang et~al.(2020{\natexlab{b}})Wang, Novikov, Zolna, Merel,
  Springenberg, Reed, Shahriari, Siegel, Gulcehre, Heess, and
  de~Freitas]{crr2020}
Ziyu Wang, Alexander Novikov, Konrad Zolna, Josh~S Merel, Jost~Tobias
  Springenberg, Scott~E Reed, Bobak Shahriari, Noah Siegel, Caglar Gulcehre,
  Nicolas Heess, and Nando de~Freitas.
\newblock {Critic Regularized Regression}.
\newblock In H.~Larochelle, M.~Ranzato, R.~Hadsell, M.~F. Balcan, and H.~Lin,
  editors, \emph{{Advances in Neural Information Processing Systems}},
  volume~33, pages 7768--7778. Curran Associates, Inc., 2020{\natexlab{b}}.

\bibitem[Williams(1992)]{reinforce1992}
Ronald~J Williams.
\newblock Simple statistical gradient-following algorithms for connectionist
  reinforcement learning.
\newblock \emph{Machine learning}, 8\penalty0 (3):\penalty0 229--256, 1992.

\bibitem[Wolf et~al.(2019)Wolf, Debut, Sanh, Chaumond, Delangue, Moi, Cistac,
  Rault, Louf, Funtowicz, et~al.]{huggingface2019}
Thomas Wolf, Lysandre Debut, Victor Sanh, Julien Chaumond, Clement Delangue,
  Anthony Moi, Pierric Cistac, Tim Rault, R{\'e}mi Louf, Morgan Funtowicz,
  et~al.
\newblock Huggingface's transformers: State-of-the-art natural language
  processing.
\newblock \emph{arXiv preprint arXiv:1910.03771}, 2019.

\bibitem[Wu et~al.(2021)Wu, Ouyang, Ziegler, Stiennon, Lowe, Leike, and
  Christiano]{wu2021recursively}
Jeff Wu, Long Ouyang, Daniel~M Ziegler, Nisan Stiennon, Ryan Lowe, Jan Leike,
  and Paul Christiano.
\newblock Recursively summarizing books with human feedback.
\newblock \emph{arXiv preprint arXiv:2109.10862}, 2021.

\bibitem[Xia et~al.(2008)Xia, Liu, Wang, Zhang, and Li]{listmle2008}
Fen Xia, Tie-Yan Liu, Jue Wang, Wensheng Zhang, and Hang Li.
\newblock Listwise approach to learning to rank: theory and algorithm.
\newblock In \emph{Proceedings of the 25th international conference on Machine
  learning}, pages 1192--1199, 2008.

\bibitem[Xu et~al.(2020)Xu, Ju, Li, Boureau, Weston, and Dinan]{xu2020recipes}
Jing Xu, Da~Ju, Margaret Li, Y-Lan Boureau, Jason Weston, and Emily Dinan.
\newblock Recipes for safety in open-domain chatbots.
\newblock \emph{arXiv preprint arXiv:2010.07079}, 2020.

\bibitem[Yang et~al.(2022{\natexlab{a}})Yang, Feng, Zhang, and
  Zhou]{sdmgan2022}
Shentao Yang, Yihao Feng, Shujian Zhang, and Mingyuan Zhou.
\newblock Regularizing a model-based policy stationary distribution to
  stabilize offline reinforcement learning.
\newblock In \emph{International Conference on Machine Learning}, pages
  24980--25006. PMLR, 2022{\natexlab{a}}.

\bibitem[Yang et~al.(2022{\natexlab{b}})Yang, Wang, Zheng, Feng, and
  Zhou]{jointmatching2022}
Shentao Yang, Zhendong Wang, Huangjie Zheng, Yihao Feng, and Mingyuan Zhou.
\newblock A regularized implicit policy for offline reinforcement learning.
\newblock \emph{arXiv preprint arXiv:2202.09673}, 2022{\natexlab{b}}.

\bibitem[Yang et~al.(2022{\natexlab{c}})Yang, Zhang, Feng, and Zhou]{wmbrl2022}
Shentao Yang, Shujian Zhang, Yihao Feng, and Mingyuan Zhou.
\newblock A unified framework for alternating offline model training and policy
  learning.
\newblock In Alice~H. Oh, Alekh Agarwal, Danielle Belgrave, and Kyunghyun Cho,
  editors, \emph{Advances in Neural Information Processing Systems},
  2022{\natexlab{c}}.

\bibitem[Yang et~al.(2024)Yang, Chen, and Zhou]{yangdense}
Shentao Yang, Tianqi Chen, and Mingyuan Zhou.
\newblock A dense reward view on aligning text-to-image diffusion with
  preference.
\newblock In \emph{Forty-first International Conference on Machine Learning},
  2024.

\bibitem[Yang et~al.(2018)Yang, Hu, Dyer, Xing, and
  Berg-Kirkpatrick]{yang2018unsupervised}
Zichao Yang, Zhiting Hu, Chris Dyer, Eric~P Xing, and Taylor Berg-Kirkpatrick.
\newblock Unsupervised text style transfer using language models as
  discriminators.
\newblock \emph{Advances in Neural Information Processing Systems}, 31, 2018.

\bibitem[Yin et~al.(2025)Yin, Yang, Xie, Yang, Sun, Awadalla, Chen, and
  Zhou]{yin2025segmentingtextlearningrewards}
Yueqin Yin, Shentao Yang, Yujia Xie, Ziyi Yang, Yuting Sun, Hany Awadalla,
  Weizhu Chen, and Mingyuan Zhou.
\newblock Segmenting text and learning their rewards for improved rlhf in
  language model, 2025.
\newblock URL \url{https://arxiv.org/abs/2501.02790}.

\bibitem[Yu et~al.(2017)Yu, Zhang, Wang, and Yu]{yu2017seqgan}
Lantao Yu, Weinan Zhang, Jun Wang, and Yong Yu.
\newblock Seqgan: Sequence generative adversarial nets with policy gradient.
\newblock In \emph{Proceedings of the AAAI conference on artificial
  intelligence}, volume~31, 2017.

\bibitem[Zhang et~al.(2019{\natexlab{a}})Zhang, Zhao, Saleh, and
  Liu]{zhang2019pegasus}
Jingqing Zhang, Yao Zhao, Mohammad Saleh, and Peter~J. Liu.
\newblock Pegasus: Pre-training with extracted gap-sentences for abstractive
  summarization, 2019{\natexlab{a}}.

\bibitem[Zhang et~al.(2021{\natexlab{a}})Zhang, Fan, Chen, and
  Zhou]{zhang2021bayesian}
Shujian Zhang, Xinjie Fan, Bo~Chen, and Mingyuan Zhou.
\newblock Bayesian attention belief networks.
\newblock In \emph{International Conference on Machine Learning}, pages
  12413--12426. PMLR, 2021{\natexlab{a}}.

\bibitem[Zhang et~al.(2021{\natexlab{b}})Zhang, Fan, Zheng, Tanwisuth, and
  Zhou]{zhang2021alignment}
Shujian Zhang, Xinjie Fan, Huangjie Zheng, Korawat Tanwisuth, and Mingyuan
  Zhou.
\newblock Alignment attention by matching key and query distributions.
\newblock \emph{Advances in Neural Information Processing Systems},
  34:\penalty0 13444--13457, 2021{\natexlab{b}}.

\bibitem[Zhang et~al.(2021{\natexlab{c}})Zhang, Gong, and
  Choi]{zhang2021knowing}
Shujian Zhang, Chengyue Gong, and Eunsol Choi.
\newblock Knowing more about questions can help: Improving calibration in
  question answering.
\newblock \emph{arXiv preprint arXiv:2106.01494}, 2021{\natexlab{c}}.

\bibitem[Zhang et~al.(2021{\natexlab{d}})Zhang, Gong, and
  Choi]{zhang2021learning}
Shujian Zhang, Chengyue Gong, and Eunsol Choi.
\newblock Learning with different amounts of annotation: From zero to many
  labels.
\newblock \emph{arXiv preprint arXiv:2109.04408}, 2021{\natexlab{d}}.

\bibitem[Zhang et~al.(2022{\natexlab{a}})Zhang, Gong, and
  Liu]{zhang2022passage}
Shujian Zhang, Chengyue Gong, and Xingchao Liu.
\newblock Passage-mask: A learnable regularization strategy for
  retriever-reader models.
\newblock \emph{arXiv preprint arXiv:2211.00915}, 2022{\natexlab{a}}.

\bibitem[Zhang et~al.(2022{\natexlab{b}})Zhang, Gong, Liu, He, Chen, and
  Zhou]{zhang2022allsh}
Shujian Zhang, Chengyue Gong, Xingchao Liu, Pengcheng He, Weizhu Chen, and
  Mingyuan Zhou.
\newblock Allsh: Active learning guided by local sensitivity and hardness.
\newblock \emph{arXiv preprint arXiv:2205.04980}, 2022{\natexlab{b}}.

\bibitem[Zhang et~al.(2022{\natexlab{c}})Zhang, Wang, Zhou, Schuurmans, and
  Gonzalez]{tempera2022}
Tianjun Zhang, Xuezhi Wang, Denny Zhou, Dale Schuurmans, and Joseph~E Gonzalez.
\newblock Tempera: Test-time prompting via reinforcement learning.
\newblock \emph{arXiv preprint arXiv:2211.11890}, 2022{\natexlab{c}}.

\bibitem[Zhang et~al.(2019{\natexlab{b}})Zhang, Kishore, Wu, Weinberger, and
  Artzi]{zhang2019bertscore}
Tianyi Zhang, Varsha Kishore, Felix Wu, Kilian~Q Weinberger, and Yoav Artzi.
\newblock Bertscore: Evaluating text generation with bert.
\newblock \emph{arXiv preprint arXiv:1904.09675}, 2019{\natexlab{b}}.

\bibitem[Zhang et~al.(2015)Zhang, Zhao, and LeCun]{zhang2015character}
Xiang Zhang, Junbo Zhao, and Yann LeCun.
\newblock Character-level convolutional networks for text classification.
\newblock \emph{Advances in neural information processing systems}, 28, 2015.

\bibitem[Zhong et~al.(2021)Zhong, Friedman, and Chen]{zhong2021factual}
Zexuan Zhong, Dan Friedman, and Danqi Chen.
\newblock Factual probing is [mask]: Learning vs. learning to recall.
\newblock \emph{arXiv preprint arXiv:2104.05240}, 2021.

\bibitem[Ziegler et~al.(2019)Ziegler, Stiennon, Wu, Brown, Radford, Amodei,
  Christiano, and Irving]{ziegler2019fine}
Daniel~M Ziegler, Nisan Stiennon, Jeffrey Wu, Tom~B Brown, Alec Radford, Dario
  Amodei, Paul Christiano, and Geoffrey Irving.
\newblock Fine-tuning language models from human preferences.
\newblock \emph{arXiv preprint arXiv:1909.08593}, 2019.

\end{thebibliography}

\clearpage
\appendix

\begin{center}
\Large
\textbf{Appendix for ``Preference-grounded Token-level Guidance for Language Model Fine-tuning''}
\end{center}

\tableofcontents

\clearpage

\section{Additional Experimental Results} \label{sec:add_results}

\subsection{Tabular Results} \label{sec:tab_results}
\begin{table}[H]
\captionsetup{font=small}
\caption{
\small
Examples of the generated discrete input-agnostic text-prompt and their classification accuracy on the corresponding test set.
} 
\label{table:prompt_example}
\centering 
\resizebox{\textwidth}{!}{
\begin{tabular}{@{}lc|lc@{}}
\toprule
\multicolumn{2}{c|}{SST-2}                       & \multicolumn{2}{c}{AG News}                      \\
 \multicolumn{1}{c}{Prompt}                                & Accuracy & \multicolumn{1}{c}{Prompt}                                & Accuracy \\ \midrule
guys filmmaker filmmaker rated Grade  & 94.18    & newsIntroduction Comments Tags Search & 85.78    \\
MovieMovieFilm rated Grade            & 94.18    & newsTopic Blog Support Category       & 85.55    \\
Rated CinemaScoreReporting Grade      & 94.01    & news RecentRecentPhotosIntroduction   & 84.53    \\
employment theater rated Oscars Grade & 93.96    & news Recent Brief LatestExample       & 84.51    \\
scene filmmaking rated comedian Grade & 93.85    & newsVirtualBlogBlogNet                & 84.33    \\ \bottomrule
\end{tabular}
}
\end{table}

\begin{table}[H]
\captionsetup{font=small}
\caption{
\small
Detailed results on CNN/DM summarization under \textbf{T5-base LM} for \Secref{sec:main_result_suma}. 
We bold the best result of each metric.
Baseline results are directly cited from RL4LMs \citep{nlpo2022}.
``Env. Reward'' denotes the environmental reward in RL4LMs.
The ``ROUGE-L'' here refers to ``Rouge-LSum'' in RL4LMs and in the Hugging Face interface, which is discussed in details in Appendix~\ref{sec:suma_details}.
In \Secref{sec:main_result_suma}, we plot the results of our method with the \textit{average} aggregation, which is the best variant in Table~\ref{table:suma_main}.
We report the mean (standard deviation) of our method over three random seeds.
} 
\label{table:suma_t5base}
\centering 
\resizebox{\textwidth}{!}{
\begin{tabular}{@{}lllllll@{}}
\toprule
Algorithm                          & Env. Reward & ROUGE-1      & ROUGE-2      & ROUGE-L  & Meteor & BertScore    \\ \midrule
Lead-3                             &             & 40.1         & 17.5         & 36.3     &   33.3 & 87.4 \\ \midrule
Supervised                         &             & 41.1         & 17.7         & 34.3    &   30.9  &  87.6 \\ \midrule
\multirow{3}{*}{PPO}               & Rouge-1     & 41.0         & 18.2         & 34.9     &  27.6  & 87.6  \\
                                   & Rouge-Avg   & 39.6         & 17.6         & 33.8     &   27.0 & 87.4 \\
                                   & Meteor      & 40.8         & 17.8         & 34.2     &   30.1 & 87.3 \\ \midrule
\multirow{3}{*}{NLPO}              & Rouge-1     & 40.4         & 18.0         & 34.4    &  27.5   & 87.5 \\
                                   & Rouge-Avg   & 40.4         & 17.7         & 34.4    &   27.4  &  87.4 \\
                                   & Meteor      & 40.5         & 18.0         & 34.3     &  29.2   & 87.2 \\ \midrule
\multirow{3}{*}{Supervised + PPO}  & Rouge-1     & 41.7         & 18.9         & 35.8      &  27.8  & 88.2 \\
                                   & Rouge-Avg   & 42.5         & 19.4         & 36.3    &   29.6   & 88.2 \\
                                   & Meteor      & 42.6         & 19.4         & 36.1    &   31.6   & 88.0 \\ \midrule
\multirow{3}{*}{Supervised + NLPO} & Rouge-1     & 42.1         & 19.3         & 36.1   &   28.7   &  88.2 \\
                                   & Rouge-Avg   & 42.4         & 19.3         & 36.3   &    29.5  &  88.2 \\
                                   & Meteor      & 42.9         & 19.4         & 36.1   &    31.9  &  88.0 \\ \midrule
Ours    (AVG)                           &             & \textbf{43.09} {\scriptsize (0.06)} & \textbf{20.17} {\scriptsize (0.04)} & \textbf{39.99} {\scriptsize (0.07)} & \textbf{35.23} {\scriptsize (0.06)} & \textbf{89.61}  {\scriptsize (0.12)} \\
Ours  (SUM)   &     & 42.86  {\scriptsize (0.08)} & 19.92  {\scriptsize (0.08)} &  39.76 {\scriptsize (0.11)}  &  34.74 {\scriptsize (0.37)} & 89.24 {\scriptsize (0.11)} \\
Ours  (MIN)    &    & 42.92  {\scriptsize (0.14)} &  20.01 {\scriptsize (0.02)} & 39.84  {\scriptsize (0.08)} &  34.88 {\scriptsize (0.13)}  &  89.33 {\scriptsize (0.07)}\\
Ours  (MAX)     &   & 42.38  {\scriptsize (0.17)} &  19.49 {\scriptsize (0.02)} & 39.34  {\scriptsize (0.09)} &  34.13 {\scriptsize (0.32)} & 89.09 {\scriptsize (0.19)} \\  
\bottomrule
\end{tabular}
}
\end{table}

\paragraph{Setup and results of the human evaluation.}
Table~\ref{table:suma_human} below presents the results of our human evaluation on CNN/DM summarization under the T5-base LM.
We generally adopt the protocol in \citet{stiennon2020learning} to evaluate the overall summary quality.
Our model is compared with the baselines Supervised, Supervised+PPO, and Supervised+NLPO in RL4LMs \citep{nlpo2022}. 
The result of the reference summaries is also presented, which is intended for sanity check rather than method comparison.
In conducting this evaluation, we randomly picked 100 articles in the test split of CNN/DM and showed to 20 qualified evaluators the summaries generated from each method, along with the article. 
The method names were anonymized. The evaluators were asked to read the article and score each summary.
Summaries are scored on a 5-Point Likert Scale \{1, 2, 3, 4, 5\}, where score 5 is the highest and 1 the lowest.
From Table~\ref{table:suma_human}, it is clear that human evaluation supports the improvements in ROUGE, Meteor, and BertScore by our method in Table~\ref{table:suma_t5base}.

\begin{table}[H]
\caption{
\small
Average human ratings on CNN/DM summarization under the T5-base LM. 
We bold the best result apart from the ground-truth Reference summary.
A detailed description on the setup is in the above text.
} 
\label{table:suma_human}
\centering 
\begin{tabular}{@{}ccccc|c@{}}
\toprule
                     & Supervised & Supervised+PPO & Supervised+NLPO & Ours & Reference \\ \midrule
Average Human Rating & 2.92       & 3.17           & 3.29            & \textbf{3.61} & 3.88      \\ \bottomrule
\end{tabular}
\end{table}

\begin{table}[H]
\captionsetup{font=small}
\caption{
Scores on each ROUGE metric for our method using sequence-level and token-level preference-based guidance in the summarization tasks in \Secref{sec:exp_abla} \textbf{(a)}.
``Seq.'' denotes our method with sequence-level preference-based guidance, and ``Token'' denotes our method with token-level preference-based guidance.
The reported numbers are mean (standard deviation) over three random seeds.
The row ``Average'' shows the average of the three ROUGE scores, \ie, (ROUGE-1 + ROUGE-2 + ROUGE-L) / 3.
} 
\label{table:seq_rew}
\centering 
\begin{tabular}{@{}ccccccc@{}}
\toprule
                             & \multicolumn{2}{c}{CNN/DM}     & \multicolumn{2}{c}{XSum}       & \multicolumn{2}{c}{CNN/DM (T5-base LM)} \\
                             & Seq. & Token & Seq. & Token & Seq.    & Token    \\ \midrule
\multicolumn{1}{c|}{ROUGE-1} & 40.20 {\scriptsize (0.07)}     & 40.94 {\scriptsize (0.02)} &  32.56 {\scriptsize (0.08)}    & 33.62 {\scriptsize (0.03)} & 42.10 {\scriptsize (0.15)}       & 43.09 {\scriptsize (0.06)}    \\
\multicolumn{1}{c|}{ROUGE-2} & 17.80 {\scriptsize (0.08)}    & 18.78 {\scriptsize (0.03)} &  ~~9.98 {\scriptsize (0.04)}    & 11.17 {\scriptsize (0.02)} & 19.23  {\scriptsize (0.11)}       & 20.17  {\scriptsize (0.04)}    \\
\multicolumn{1}{c|}{ROUGE-L} & 37.08  {\scriptsize (0.06)}    & 38.17  {\scriptsize (0.03)} &  25.11 {\scriptsize (0.07)}     &  26.33 {\scriptsize (0.05)} & 38.09  {\scriptsize (0.14)}       &  39.99 {\scriptsize (0.07)}   \\ \midrule
\multicolumn{1}{c|}{Average} &    31.69       &   32.63     &  22.55         &   23.71      &  33.14 & 34.42  \\ \bottomrule
\end{tabular}
\end{table}

\begin{table}[H]
\captionsetup{font=small}
\caption{
Scores on each ROUGE metric for our method with and without the reward-function retraining scheme in the summarization tasks in \Secref{sec:exp_abla} \textbf{(b)}.
``Without Retrain'' denotes our method without reward-function retraining, and ``With Retrain'' denotes our method with reward-function retraining.
The reported numbers are mean (standard deviation) over three random seeds.
The row ``Average'' shows the average of the three ROUGE scores, \ie, (ROUGE-1 + ROUGE-2 + ROUGE-L) / 3.
} 
\label{table:wo_rew_retrain}
\centering 
\resizebox{1.0\textwidth}{!}{
\begin{tabular}{@{}ccccccc@{}}
\toprule
                             & \multicolumn{2}{c}{CNN/DM}     & \multicolumn{2}{c}{XSum}       & \multicolumn{2}{c}{CNN/DM (T5-base LM)} \\
                             & Without Retrain & With Retrain & Without Retrain & With Retrain & Without Retrain    & With Retrain    \\ \midrule
\multicolumn{1}{c|}{ROUGE-1} & 40.83 {\scriptsize (0.10)}     & 40.94 {\scriptsize (0.02)} & 33.45 {\scriptsize (0.11)}    & 33.62 {\scriptsize (0.03)} & 42.98 {\scriptsize (0.08)}       & 43.09 {\scriptsize (0.06)}    \\
\multicolumn{1}{c|}{ROUGE-2} & 18.70 {\scriptsize (0.07)}    & 18.78 {\scriptsize (0.03)} & 11.07 {\scriptsize (0.06)}    & 11.17 {\scriptsize (0.02)} & 20.09 {\scriptsize (0.06)}       & 20.17 {\scriptsize (0.04)}    \\
\multicolumn{1}{c|}{ROUGE-L} & 38.07 {\scriptsize (0.09)}    & 38.17 {\scriptsize (0.03)} & 26.23 {\scriptsize (0.10)}     & 26.33 {\scriptsize (0.05)} & 39.87 {\scriptsize (0.08)}       & 39.99 {\scriptsize (0.07)}    \\ \midrule
\multicolumn{1}{c|}{Average} & 32.53           & 32.63        & 23.58           & 23.71        & 34.31              & 34.42           \\ \bottomrule
\end{tabular}
}
\end{table}

\begin{table}[H]
\captionsetup{font=small}
\caption{
Scores on each ROUGE metric for the summarization task on CNN/DM in \Secref{sec:exp_abla} \textbf{(c)}, where we vary the number of sequences used to learn the token-level guidance.
The reported numbers are mean (standard deviation) over three random seeds.
The row ``Average'' shows the average of the three ROUGE scores, \ie, (ROUGE-1 + ROUGE-2 + ROUGE-L) / 3.
} 
\label{table:traj_cnn}
\centering 
\begin{tabular}{@{}cccccc@{}}
\toprule
                             & \multicolumn{5}{c}{Number of Sequences}                                  \\ \cmidrule(l){2-6} 
                             & 2            & 3            & 5              & 7            & 9             \\ \midrule
\multicolumn{1}{c|}{ROUGE-1} & 40.80 {\scriptsize (0.06)} & 40.94 {\scriptsize (0.02)} & 40.87 {\scriptsize (0.09)}   & 40.86 {\scriptsize (0.08)} & 40.95 {\scriptsize (0.01)}  \\
\multicolumn{1}{c|}{ROUGE-2} & 18.70 {\scriptsize (0.04)} & 18.78 {\scriptsize (0.03)} & 18.71 {\scriptsize (0.02)}   & 18.74 {\scriptsize (0.06)}  &  18.78 {\scriptsize (0.01)}  \\
\multicolumn{1}{c|}{ROUGE-L} & 38.05 {\scriptsize (0.03)} & 38.17 {\scriptsize (0.03)}  & 38.09  {\scriptsize (0.07)} & 38.08 {\scriptsize (0.08)} & 38.18 {\scriptsize (0.02)} \\ \midrule
\multicolumn{1}{c|}{Average} &   32.52      &    32.63     &      32.56     &   32.56      &    32.64      \\ \bottomrule
\end{tabular}

\end{table}

\begin{table}[H]
\captionsetup{font=small}
\caption{
Scores on each ROUGE metric for the summarization task on XSum in \Secref{sec:exp_abla} \textbf{(c)}, where we vary the number of sequences used to learn the token-level guidance.
The reported numbers are mean (standard deviation) over three random seeds.
The row ``Average'' shows the average of the three ROUGE scores, \ie, (ROUGE-1 + ROUGE-2 + ROUGE-L) / 3.
} 
\label{table:traj_xsum}
\centering 
\begin{tabular}{@{}cccccc@{}}
\toprule
                             & \multicolumn{5}{c}{Number of Sequences}                               \\ \cmidrule (l){2-6} 
                             & 2            & 3            & 5            & 7            & 9            \\ \midrule
\multicolumn{1}{c|}{ROUGE-1} & 33.54  {\scriptsize (0.06)} & 33.62 {\scriptsize (0.03)} & 33.56 {\scriptsize (0.08)} & 33.56 {\scriptsize (0.02)}  & 33.63 {\scriptsize (0.02)} \\
\multicolumn{1}{c|}{ROUGE-2} & 11.12 {\scriptsize (0.04)} & 11.17 {\scriptsize (0.02)} & 11.12 {\scriptsize (0.05)} & 11.19 {\scriptsize (0.05)} & 11.20 {\scriptsize (0.03)} \\
\multicolumn{1}{c|}{ROUGE-L} & 26.26 {\scriptsize (0.06)} & 26.33 {\scriptsize (0.05)} & 26.28 {\scriptsize (0.06)} & 26.34 {\scriptsize (0.06)} & 26.36 {\scriptsize (0.03)} \\ \midrule
\multicolumn{1}{c|}{Average} &    23.64     &   23.71      &   23.65      &   23.70      & 23.73         \\ \bottomrule
\end{tabular}
\end{table}

\subsection{Further Ablation Study} \label{sec:more_abla}
In learning the preference-based \textit{sequence-level} guidance in \Secref{sec:exp_abla}, the aggregation function $f(\cdot)$ in \Secref{sec:method_agg_func} is removed, since it is inapplicable and unnecessary to the sequence-level reward function.
For the minimalist LM training objectives Eqs.~\eqref{eq:prompt_pol_objective} and~\eqref{eq:sum_pol_objective} in \Secref{sec:policy_training}, we change them to the corresponding versions that use sequence-level guidance.
Self-normalization  in reward-weighted MLE Eq.~~\eqref{eq:sum_pol_objective} is removed, since it is again inapplicable and unnecessary to the sequence-level setting.

In this section, we continue our discussion in the Ablation Study (\Secref{sec:exp_abla}) by answering the following additional questions on our method.



\textbf{(a):} \textit{Is our method robust to the hyperparameter(s): temperature $\beta$ and balancing coefficient $\alpha$?}

\begin{figure*}[tb]
     \centering
     \begin{subfigure}[b]{0.33\textwidth}
         \centering
         \includegraphics[width=\textwidth]{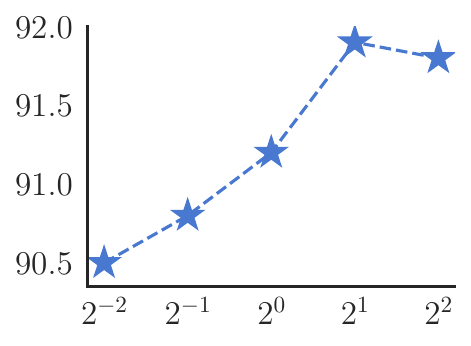}
         \captionsetup{font=small}
         \caption{\small{Temperature $\beta$ (SST-2)}}
         \label{fig:temp_sst_2}
     \end{subfigure}
     \hfill
     \begin{subfigure}[b]{0.32\textwidth}
         \centering
         \includegraphics[width=\textwidth]{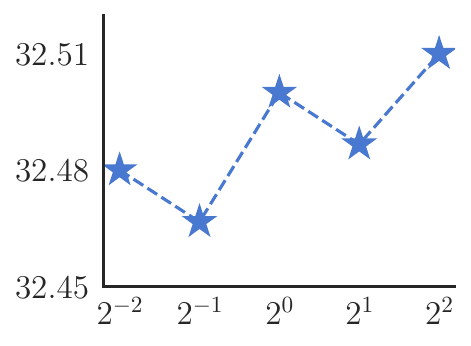}
         \captionsetup{font=small}
         \caption{\small{Temperature $\beta$ (CNN/DM)}}
         \label{fig:temp_cnn}
     \end{subfigure}
     \hfill
    \begin{subfigure}[b]{0.33\textwidth}
         \centering
         \includegraphics[width=\textwidth]{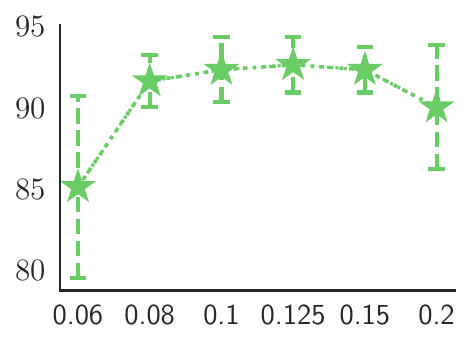}
         \captionsetup{font=small}
         \caption{\small{Balancing Coeff. $\alpha$ (SST-2)}}
         \label{fig:entropy_sst_2}
     \end{subfigure}
     \captionsetup{font=small}
        \caption{ 
        Line plots comparing the performance under different values of the hyperparameter $\beta$ in Eq.~\eqref{eq:max_agg} and $\alpha$ in Eq.~\eqref{eq:prompt_pol_objective}. 
        The plotted numbers are mean over three random seeds.
        Error bars show one standard deviation.
        }
        \label{fig:vary_param}
        \vspace{5mm}
\end{figure*}

To study the choice of the temperature parameter $\beta$ in the soft-maximum/minimum aggregation Eq.~\eqref{eq:max_agg}, we vary the value of $\beta$ in the MIN variant in Tables~\ref{table:prompt_main} and \ref{table:suma_main} from $\beta=2$.
Furthermore, to study the balancing coefficient $\alpha$ in the REINFORCE-style LM-training approach Eq.~\eqref{eq:prompt_pol_objective}, we vary the $\alpha$ parameter in the AVG variant in Table~\ref{table:prompt_main} from $\alpha=2^{-3}$.
Fig.~\ref{fig:vary_param} respectively shows the prompt results on the SST-2 dataset and  the summarization results on the CNN/DM dataset.
For summarization, we again plot the average ROUGE scores, with the breakdown scores of the three ROUGE metrics in Table~\ref{table:temp_cnn} below.

Recall that the best baseline result on SST-2 in Table~\ref{table:prompt_main} is $90.5$, and on CNN/DM in Table~\ref{table:suma_main} is $31.3$.
We see that our method can achieve competitive results on a relatively wide range of the temperature $\beta$.
A too-small value of $\beta$, such as $0.25$ and $0.5$, may incur a harder optimization problem and thus an inferior performance on both prompt and summarization tasks.

For the choice of the balancing coefficient $\alpha$,
we see that our method provides competitive results in a relatively wide range of $\alpha \in [0.08, 0.15]$, when compared to the best baseline result of $90.5$ in Table~\ref{table:prompt_main}.
A too-small value of $\alpha$ may not prevent the REINFORCE-style method from pre-mature convergence.
The resulting LM therefore may not sufficiently explore the sampling space or capture multiple good behavior-modes, resulting in an inferior and highly varying performance.
A too-large value of $\alpha$ distracts the optimization of the LM, and again leads to a worse result.

\begin{table}[H]
\vspace{-1em}
\captionsetup{font=small}
\caption{
Scores on each ROUGE metric for the summarization task on CNN/DM, where we vary the temperature parameter $\beta$ in the \textit{soft-minimum} aggregation Eq.~\eqref{eq:max_agg}.
The reported numbers are mean (standard deviation) over three random seeds.
The row ``Average'' shows the average of the three ROUGE scores, \ie, (ROUGE-1 + ROUGE-2 + ROUGE-L) / 3.
} 
\label{table:temp_cnn}
\centering 
\begin{tabular}{@{}cccccc@{}}
\toprule
                             & $\beta=2^{-2}$ & $\beta=2^{-1}$ & \multicolumn{1}{l}{$\beta=2^0$} & \multicolumn{1}{l}{$\beta=2^1$} & \multicolumn{1}{l}{$\beta=2^2$} \\ \midrule
\multicolumn{1}{c|}{ROUGE-1} & 40.77 {\scriptsize (0.11)}   & 40.74 {\scriptsize (0.09)}   &  40.79 {\scriptsize (0.11)}                    & 40.78 {\scriptsize (0.06)}                    & 40.80 {\scriptsize (0.01)}                    \\
\multicolumn{1}{c|}{ROUGE-2} & 18.67 {\scriptsize (0.06)}   & 18.68 {\scriptsize (0.05)}   &  18.68 {\scriptsize (0.09)}                     & 18.67 {\scriptsize (0.03)}                    &  18.71 {\scriptsize (0.04)}                    \\
\multicolumn{1}{c|}{ROUGE-L} & 38.00 {\scriptsize (0.10)}    & 37.98 {\scriptsize (0.08)}   & 38.03 {\scriptsize (0.12)}                    & 38.01 {\scriptsize (0.04)}                    & 38.02 {\scriptsize (0.01)}                    \\\midrule
\multicolumn{1}{c|}{Average} &    32.48       &    32.47       &     32.50                     &  32.49                         & 32.51                           \\ \bottomrule
\end{tabular}
\vspace{-1em}
\end{table}

\textbf{(b):} \textit{How does our method perform in generating longer prompts compared with the baseline?}

To further validate the harm of the delayed-feedback issue to the related LM-training methods that learn under the sparse sequence-level feedback, we compare our method with RLPrompt \citep{rlprompt2022} on generating prompts with length increased from $5$ to $10$ and to $20$ tokens, on the SST-2 dataset. 
Table~\ref{table:longer_prompt} below shows the results.

\begin{table}[H]
\vspace{-1em}
\captionsetup{font=small}
\caption{
\small
Test accuracy on the prompt task on the SST-2 dataset,
for our method and RLPrompt on generating prompts with a length of $5$, $10$, and $20$ tokens.
We report the mean and standard deviation over three random seeds.
} 
\label{table:longer_prompt}
\centering 
\begin{tabular}{@{}rccc@{}}
\toprule
          & RLPrompt   & Ours (AVG)  & Performance Gap \\ \midrule
5 Tokens  & 90.5 {\scriptsize (1.5)} & 92.6 {\scriptsize (1.7)} & 2.1             \\
10 Tokens & 75.8 {\scriptsize (7.6)} & 86.0 {\scriptsize (2.9)} & 10.2            \\
20 Tokens & 65.2 {\scriptsize (6.0)} & 80.9 {\scriptsize (4.5)} & 15.7            \\ \bottomrule
\end{tabular}
\vspace{-1.5em}
\end{table}

We see that RLPrompt performs worse than our method on generating longer prompts.
In particular, the performance gap increases as the prompt length (feedback delaying) increases. 
This comparison can further demonstrate the harm of the delayed-feedback issue in training text-generation LMs, and that our framework, in particular our preference-grounded token-level guidance for LM training, is a viable solution to it.

It is intrigued that the results of both methods deteriorate with the prompt length. 
After checking the generated prompts from our method, we find that longer prompts mostly contain many repeated tokens, as shown by the following example prompt of length $20$
\begin{center}
    \begin{verbatim}
PerformanceExceptionMovieMovieMovieMovieMovieMovieMovieVideoVideoVideoVideo\
VideoVideoVideoImageVideoImageImage
    \end{verbatim}
\end{center}
which is separated into two lines at the location of ``\texttt{\textbackslash}'' due to the page-width limit.
In this prompt example, the tokens \texttt{Movie} and \texttt{Video} are each consecutively repeated  seven times, and the bi-gram \texttt{ImageVideo} is repeated two times. 
Such prompts with heavy repetitions may confuse the downstream classifier.\footnote{A detailed description of the prompt task is deferred to Appendix~\ref{sec:details_prompt_tasks}.} 
This aligns with our intuition that a clear and succinct instruction is preferable than a long but verbose one.

As a side note, in generating Table~\ref{table:longer_prompt}, we use the default hyperparameters for both our method and RLPrompt. 
It is possible that RLPrompt requires careful tuning for generating longer prompts, due to the delayed-feedback issue that we try to address. 
We leave a thorough tuning of RLPrompt on long-prompt generation as a future work.

\textbf{(c):} \textit{Is the efficacy of our framework tied to the  specific preference sources considered in \Secref{sec:exp}?}

To investigate whether the performance of our framework is tied to the specific preference-sources considered in the experiment section (\Secref{sec:exp}), inspired by RL4LMs \citep{nlpo2022}, we simulate the sequence-level preference on the summarization task by using another two automatic metrics ``Rouge-avg'' and ``Rouge-avg2'', rather than the classical Meteor score \citep{meteor2005} in \Secref{sec:exp}.
Table~\ref{table:diff_env_rew} below presents the ROUGE scores of our method under each of the three preference sources  on the CNN/DM dataset under the T5-base LM.
For a more thorough investigation, we provide the results for our method both with and without the guidance re-estimation scheme.
The baseline results in Table~\ref{table:diff_env_rew} below come from the best baseline method in Table~\ref{table:suma_t5base} of Appendix~\ref{sec:tab_results}.

\begin{table}[H]
\vspace{-1em}
\captionsetup{font=small}
\caption{
Results for our method on CNN/DM summarization under T5-base LM  when using different automatic metrics to simulate the sequence-level preference.
We provide the detailed ROUGE scores for our method both with and without guidance re-estimation.
``Baseline'' denotes the results of the best baseline method in Table~\ref{table:suma_t5base} of Appendix~\ref{sec:tab_results}.
The reported numbers are the mean over three random seeds.
The row ``Average'' shows the average of the three ROUGE scores, \ie, (ROUGE-1 + ROUGE-2 + ROUGE-L) / 3.
} 
\label{table:diff_env_rew}
\centering 
\resizebox{\textwidth}{!}{
\begin{tabular}{@{}cccccccc@{}}
\toprule
                        & \multirow{2}{*}{Baseline}     & \multicolumn{3}{c}{With Guidance Re-estimation}     & \multicolumn{3}{c}{Without Guidance Re-estimation}       \\
                        &     & Rouge-avg & Rouge-avg2 & Meteor & Rouge-avg & Rouge-avg2 & Meteor    \\ \midrule
\multicolumn{1}{c|}{ROUGE-1} &  42.9  &  43.14   & 43.07 &  43.09   &  42.96 &    42.98    & 42.98    \\
\multicolumn{1}{c|}{ROUGE-2} &  19.4  &  20.18   & 20.12  &  20.17    &   20.07   &   20.05   &  20.09   \\
\multicolumn{1}{c|}{ROUGE-L} &  36.1  &  39.93   &   39.89   &   39.99      &  39.80   &    39.77       &   39.87   \\ \midrule
\multicolumn{1}{c|}{Average} &  32.8  &   34.42      &   34.36     &    34.42       &    34.28      &  34.27 & 34.31 \\ \bottomrule
\end{tabular}
}
\end{table}

Concretely, these two new automatic metrics ``Rouge-avg'' and ``Rouge-avg2'' are constructed as 
\begin{equation*}
    \begin{split}
        \text{Rouge-avg} &= 0.5 \times \text{ROUGE-1} + 0.5 \times \text{ROUGE-2} + 0.5 \times \text{ROUGE-L} \,, \\
        \text{Rouge-avg2} &= 0.5 \times \text{ROUGE-1} + 0.5 \times 2 \times \text{ROUGE-2} + 0.5 \times \text{ROUGE-L}\,,
    \end{split}
\end{equation*}
where the ``Rouge-avg'' metric is exactly the same as that in the RL4LMs \citep{nlpo2022}.
The ``Rouge-avg2'' metric is constructed by multiplying ROUGE-2 by $2$ to make its numerical value similar to the others.

It is clear that changing the preference source from Meteor to these two alternative metrics does not significantly alter the performance of our method, especially when compared to the performance improvement of our method over the best baseline method in Table~\ref{table:suma_t5base} of Appendix~\ref{sec:tab_results}.
This set of comparisons confirms that the efficacy of our framework is generally not tied to a specific preference source.
It could also further corroborate the effectiveness of our preference-grounding perspective on guiding the LM training.

\section{Additional Experiment Details} \label{sec:exp_details}

\subsection{Prompt Generation} \label{sec:prompt_details}

\myparagraph{Implementation Details.} 
To ensure a fair comparison, the implementation of our framework is based on the official codebase of RLPrompt available at \href{https://github.com/mingkaid/rl-prompt}{https://github.com/mingkaid/rl-prompt},
and the Hugging Face library \citep{huggingface2019}.
We have  provided some implementation details in \Secref{sec:main_result_prompt}.
Here we continue the discussion.

The LM $\pi_\theta$ is parametrized as a frozen distilGPT-2 model with parameter $\theta$ being one MLP-layer of size $2048$ inserted right before the output head.
The token-level reward function $r_\phi$ is implemented as a distilGPT-2 with a two-layer projection-MLP of sizes $2048$ and $1$ on top.
The LM $\pi_\theta$ is trained by a maximum of $12000$ steps, with early stopping based on the validation set.
The reward training is reconducted every $1000$ steps during the first $6000$ steps of the LM training process
and is (and almost always) early stopped.
RoBERTa-large is used \citep{liu2019roberta} as the pre-trained downstream LM $\pi_\dlm$.

\myparagraph{Datasets.}
We use the standard datasets provided in the RLPrompt codebase \citep{rlprompt2022}.
We test on three popular few-shot classification datasets in prior work \citep[\eg,][]{gao2020making,sun2022black}, \ie, two sentiment binary-classification datasets SST-2 \citep{socher2013recursive} and Yelp Polarity \citep{zhang2015character}, and the topic four-way-classification dataset AG News \citep{zhang2015character}.
In keeping with the standard few-shot setting \citep{perez2021true}, both the training and the validation sets have $16$ examples per class.
To mitigate the randomness in the few-shot setting, each dataset is subsampled into five few-shot training and validation sets, while the test set is standard.
We train our models on each few-shot (sub-)dataset with three random seeds and evaluate three generated prompts in each case.
For all three tested datasets, we report the average test accuracy and standard deviation across all evaluated prompts in all random seeds and all few-shot (sub-)datasets.

\myparagraph{Hyperparameters.} Apart from the hyperparameters discussed in the ablation study (\Secref{sec:exp_abla} and Appendix~\ref{sec:more_abla}), most other hyperparameters as well as the training and evaluation procedures of our framework follow RLPrompt.
Additionally, we list the important hyperparameters for training our reward model in Table~\ref{table:prompt_param_rew}, and important hyperparameters for training our LM in Table~\ref{table:prompt_param_policy}.
The generated prompts have a fixed length of $5$.
The same hyperparameters are used in all tested datasets.

\myparagraph{Baselines.} For the baseline results in Table~\ref{table:prompt_main}, we rerun the codebase of RLPrompt under the same random seeds and evaluation script as our method.
Other baseline results are from the literature \citep{rlprompt2022,tempera2022}.
We note that our reported RLPrompt results have some small discrepancies compared to the original paper's results. 
We have confirmed our reproduced results with RLPrompt's authors and with Table 2 of the recent TEMPERA paper \citep{tempera2022}.

\begin{tabular}{cc}
    \begin{minipage}{.5\linewidth}
\begin{table}[H]
\captionsetup{font=small}
\caption{
\small Hyperparameters for training our reward model in the prompt-generation task.
} 
\label{table:prompt_param_rew} 
\centering 
\begin{tabular}{@{}ll@{}}
\toprule
Hyperparameter              & Value                                       \\ \midrule
Gradient clipping norm      &          5.0                               \\
Max train steps & 10000 \\
Steps per epoch & 100 \\
Number of epochs & 100 \\
Learning rate               &         5e-5                               \\
Batch size                  &                 64                           \\
Learning-rate decay         &                 0.8                         \\
Learning-rate scheduler     & \texttt{ReduceLROnPlateau} \\
Scheduler patience          &                2                            \\
Early-stop count            &                   7                         \\
Optimizer & Adam \citep{adam2014} \\
Backbone                    &              distilGPT-2                      \\ 
& \\
\bottomrule
\end{tabular}
\end{table}
\end{minipage} &
\begin{minipage}{.5\linewidth}
\begin{table}[H]
\captionsetup{font=small}
\caption{
\small Hyperparameters for training our LM in the prompt-generation task.
} 
\label{table:prompt_param_policy} 
\centering 
\begin{tabular}{@{}ll@{}}
\toprule
Hyperparameter              & Value                       \\ \midrule
Gradient clipping norm      &         5.0                 \\
Max train steps & 12000 \\
Steps per epoch & 500 \\
Number of epochs & 24  \\
Learning rate               &      5e-5                   \\
Batch size                  &    32                        \\
Learning-rate decay         &          0.8              \\
Learning-rate scheduler     & \texttt{ReduceLROnPlateau}                    \\
Scheduler patience          &             2                               \\
Early-stop count            &                 7           \\
Optimizer & Adam  \\
Backbone                    &        distilGPT-2       \\ 
Reward retrain period & 1000 steps \\
\bottomrule
\end{tabular}
\end{table}
\end{minipage} 
\end{tabular}

\subsection{Text Summarization} \label{sec:suma_details}

\myparagraph{Implementation Details and Hyperparameters.} 
The implementation of our framework is based on the Hugging Face library \citep{huggingface2019}.
We have provided some implementation details in \Secref{sec:main_result_suma}.
The discussion is continued here.

Due to our limited computational resources, unless explicitly mentioned, we use the standard T5-small model \citep{t52020} for the LM.
Similar to the prompt tasks, the token-level reward function is implemented also as a T5-small model, with a two-layer projection-MLP on top with sizes $2048$ and $1$.
The LM $\pi_\theta$ is trained for a standard $5$ epochs.
Apart from the hyperparameters discussed in the ablation study (\Secref{sec:exp_abla} and Appendix~\ref{sec:more_abla}), most other hyperparameters as well as the training and evaluation procedure of our framework follow the standard setting of using a T5 model for text summarization on the Hugging Face library.
Additionally, we list the important hyperparameters for training our reward model in Table~\ref{table:suma_param_rew}, and important hyperparameters for training our LM in Table~\ref{table:suma_param_policy}.
The same hyperparameters are used in both the CNN/DailyMail and the XSum datasets.

We note that the ROUGE-L metric we report is technically the \texttt{rougeLsum} metric from the Hugging Face interface and in the RL4LMs' codebase \citep{nlpo2022}. 
This one matches the result scales in prior work especially on texts with newlines (``\texttt{\textbackslash n}''), as reported in this \href{https://github.com/huggingface/datasets/issues/617#issuecomment-691615081}{GitHub issue}.

\myparagraph{Baselines.} For the baseline methods' results in Table~\ref{table:suma_main}, we rerun the codebase of RL4LMs \citep{nlpo2022} with a T5-small model as our method.
We have carefully tuned the (supervised+) PPO/NLPO in RL4LMs on several hyperparameters, such as \texttt{learning\_rate}, \texttt{kl\_div:coeff}, \texttt{kl\_div:target\_kl}, and so on.
Furthermore, we ran these baseline methods on the same random seeds as our method and we provide error bars.
Since we use the T5-small model and the same random seeds for both our method and the baselines, our reported results are therefore (more) fair comparisons.

\begin{tabular}{cc}
    \begin{minipage}{.5\linewidth}
\begin{table}[H]
\captionsetup{font=small}
\caption{
\small Hyperparameters for training our reward model in the text-summarization task.
} 
\label{table:suma_param_rew} 
\centering 
\begin{tabular}{@{}ll@{}}
\toprule
Hyperparameter              & Value                                       \\ \midrule
Gradient clipping norm      &          5.0                              \\
Number of epochs &  1 \\
Amount of training data & $10\%$ of training set \\
Learning rate               &        5e-5                                \\
Batch size                  &            32                                \\
Optimizer & Adam  \\
Backbone                    &             T5-small                       \\ 
& \\
\bottomrule
\end{tabular}
\end{table}
\end{minipage} &
\begin{minipage}{.42\linewidth}
\begin{table}[H]
\captionsetup{font=small}
\caption{
\small Hyperparameters for training our LM in the text-summarization task.
} 
\label{table:suma_param_policy} 
\centering 
\resizebox{\textwidth}{!}{
\begin{tabular}{@{}ll@{}}
\toprule
Hyperparameter              & Value                       \\ \midrule
Gradient clipping norm      &        5.0                  \\
Number of epochs &  5 \\
Learning rate               &         5e-5                \\
Batch size                  &            32                \\
Optimizer & AdamW \citep{adamw2017}   \\
Weight decay & 0.0 \\
Backbone                    &        T5-small       \\ 
Reward retrain period & 0.5 epoch \\
\bottomrule
\end{tabular}
}
\end{table}
\end{minipage} 
\end{tabular}

\section{A Na\"ive Numeric Example for the \textit{Average} Aggregation} \label{sec:avg_agg}

This section provides a na\"ive numeric comparison that
the \textit{average} aggregation in \Secref{sec:method_agg_func} will not automatically favor longer sequences, while the classical \textit{summation} will. 

Suppose we have $K=2$ sequences $\tau^1$ and $\tau^2$ for preference learning, respectively having length $T^1=5$ and $T^2=15$. For simplicity, assume that all tokens in $\tau^1$ and $\tau^2$ are the same and all have reward $1$, \ie, $r_{\phi}(s^k_t, a^k_t) = 1, \forall\, k, t$. 
The average sequence length $C$ is then $C = (1/2) \times (5+15) = 10$.
For the first sequence $\tau^1$, the \textit{average}-aggregated sequence-level evaluation $e^{\mathrm{avg}}_\phi(\tau^1) = (10/5) \times \sum_{t=0}^4 1 = (10/5) \times 5 = 10$. And for the second sequence $\tau^2$, $e^{\mathrm{avg}}_\phi(\tau^2) = (10/15) \times \sum_{t=0}^{14} 1 =  (10/15) \times 15 = 10$. Therefore, no sequence will be automatically preferred based only on the length.

By contrast, when using the classical \textit{summation} as the aggregation function, $\tau^1$ will be evaluated as $\sum_{t=0}^4 1 = 5$ while $\tau^2$ will be evaluated as $ \sum_{t=0}^{14} 1 = 15$. So, indeed, the longer sequence $\tau^2$ will be automatically preferred.

\section{Details on the Prompt Generation Task} \label{sec:details_prompt_tasks}

\myparagraph{Task Description.}
In discrete text-prompt generation \citep[\eg,][]{gpt32020,an2022input}, we input a discrete text-prompt $\va$ and an observation sequence $o$ to a large pre-trained downstream LM $\pi_\dlm(y_\dlm \given \va, o)$ to directly classify text $o$, without finetuning $\pi_\dlm$.
Here, $y_\dlm$ denotes the output of the large downstream LM $\pi_\dlm$ on the observation text $o$ prompted by text $\va$.
We follow the classical prompt setting \citep[\eg,][]{gpt32020,schick2020s,rlprompt2022} that solves the classification problem by an encoder-only downstream LM via token infilling.
Classification is reduced to selecting tokens corresponding to some predefined class labels, known as verbalizers, such as ``happy'' for positive and ``sad'' for negative.
The set of verbalizers is denoted as~$\gC$.
As an example, to classify an observation text $o$ by prompt $\va$ using an encoder-only downstream LM $\pi_\dlm$, we input a template such as ``\texttt{[o] [a] [MASK]}'' to $\pi_\dlm$, and select the most probable verbalizer token that fills into \texttt{[MASK]}.

\myparagraph{Setting.}
In our input-agnostic setting, the generated prompt is independent of the observation text $o$.
During inference time, only the learned prompts are used and the LM $\pi_\theta$ is discarded.
The initial input $x$ to $\pi_\theta$ is a dummy, and the target $y$ is the class label in the mask position.
We also adopt the few-shot setting, where the training set consists of a small number of samples per class.
There is a larger standard test set for evaluation.
With a fixed length $T$, the goal is to find discrete text-prompts $\va = (a_0, \ldots, a_{T-1})$ that have high test accuracy.

\myparagraph{Source of the Preference.}
For learning the token-level guidance, we simulate the sequence-level preference by the recently proposed stepwise metric $\gR_{\mathrm{step}}$ in \citet{rlprompt2022}, \ie, the higher the metric value the better prompt.
This choice ensures a fair comparison with RLPrompt \citep{rlprompt2022} and avoids a potential overfitting that we train and evaluate the LM on the same
evaluation metric ``accuracy''.

Given a prompt $\va$, observation text $o$, and the true class label $y \in \gC$, $\gR_{\mathrm{step}}$ measures the gap between the true class's probability and the highest probability in other classes.
The gap is defined as 
\begin{equation*}
    \mathrm{Gap}_o(\va, y) = \pi_\dlm(y\given \va, o) - \max_{y'\in \gC, y' \ne y} \pi_\dlm(y'\given \va, o),
\end{equation*}
where $\mathrm{Gap}_o(\va, y) > 0$ when the prediction $y_\dlm(\va,o)$ for text $o$ is correct and $< 0$ otherwise.
Define the indicator for correct prediction for $o$, $\mathrm{Corr}_o$, as $\mathrm{Corr}_o = \vone\{\mathrm{Gap}_o(\va, y) > 0 \}$.
The stepwise metric $\gR_{\mathrm{step}}$ for prompt $\va$ on observation text $o$ and true class label $y$ is define as
\begin{equation*}
    \gR_{\mathrm{step}}(y_\dlm(\va,o), y) = \lambda_1^{1-\mathrm{Corr}_o} \lambda_2^{\mathrm{Corr}_o} \times  \mathrm{Gap}_o(\va, y),
\end{equation*}
where $\lambda_1 = 180$ and $\lambda_2 = 200$.
In the experiments (\Secref{sec:exp} and Appendix~\ref{sec:more_abla}), we report test accuracy as in prior works.

\myparagraph{LM Training.}
Since the prompt-generation task does not assume the availability of supervised data --- the ground-truth prompts, the LM $\pi_\theta$ is trained by the REINFORCE-style update in \Secref{sec:policy_training_without_data} to automatically discover highly-accurate prompts.

\section{More Related Work} \label{sec:more_related}
\myparagraph{Prompt Generation.}
Prior works \citep[\eg,][]{gpt2,gpt32020,schick2020exploiting,sanh2021multitask} have shown that manual prompts can steer LMs to perform NLP tasks
in the few/zero-shot setting.
In general, prompts can be discrete, consisting of real token-strings; or can be continuous, where the prompts are entirely free word-embeddings that do not map to real tokens. 
Several works \citep[\eg,][]{qin2021learning,prefixtuning2021,hambardzumyan2021warp,liu2021gpt,zhong2021factual,lester2021power} tune continuous soft prompts using gradient descent, which typically requires some expensive gradient information \citep{sun2022black,diao2022black}.
In this work, we apply our framework to the task of input-agnostic discrete-prompt optimization due to its challenging setting, better human understandability of the learned prompts \citep{khashabi2021prompt,promptsurvey2021}, potential transferability across LMs \citep{su2021transferability,perez2021true,rlprompt2022},
and more robustness in the low-data regime \citep{prefixtuning2021}.
Recent works propose some new settings such as input-dependent prompt-tuning \citep{tempera2022}, which are potential further applications of our framework and are left for future work.

\myparagraph{Text Summarization.}
Apart from using RL techniques discussed in Sections~\ref{sec:related_work}, prior works on text summarization \citep[\eg,][]{bart2019,t52020,zhang2019pegasus,longt52021,mt52021} mainly focus on structural designs of the LMs and improvements on the source of the (pre-)training data, where the LMs are typically trained by vanilla MLE on the supervised data.
In this paper, we apply our preferenced-grounded token-level guidance to this task by considering a weighted-MLE objective for LM training.
The weights given by the learned reward function reflect some sequence-level preference among multiple candidate summaries.
Our framework thus has the potential to learn and improve from lower-quality data, and generate summaries fulfilling more general evaluation metrics, such as human preference. 

\myparagraph{Weighted MLE in NLP.}
Though not a very common techinique, the approach of weighted MLE has been adopted in prior NLP research.
For example, RAML \citep{norouzi2016reward} samples outputs proportionally to its exponentiated scaled ``reward" (negative edit/Hamming distance) using stratified sampling.
GOLD \citep{pang2020text} frames text generation as an offline RL problem with expert demos and learns from the demos by importance weighting, where training examples with higher probability under the model are weighted higher.
Besides, \citet{ghosh2021helpful} apply the weighted MLE technique to table-to-text generation and \citet{Junczys2018approaching} apply this technique to grammatical error correction for machine translation.
Our token-level reward-weighted MLE in \Secref{sec:policy_training_with_data} adds to this research thread by emphasizing the important tokens in the supervised sequences and downweighting the unimportant tokens. 
This design may better utilize the LM capacity and the optimization budget.
The efficacy of our reward-weighted MLE is experimentally verified in \Secref{sec:main_result_suma}.

\myparagraph{Align LMs with Preference.}
Similar to our paper, prior works on aligning LMs with preference typically focus on adjusting the pretrained LMs,  where preference comes from human feedback or from some automatic metrics.
A classical strategy is to add external filters on top of the pretrained LMs for the generated text sequences or for the training sequences  \citep[\eg,][]{xu2020recipes}, where the LMs are trained using MLE on abundant supervised data.
Another classical approach finetunes LMs using supervised learning (vanilla MLE) on some curated/improved datasets \citep{hancock2019learning,solaiman2021process,scheurer2022training}, or on massive highly-curated collections of tasks phrased as instructions for supervised finetuning the LMs
\citep{zhang2021knowing, zhang2021learning, sanh2022multitask,chung2022scaling}.
Apart from supervised learning, reinforcement learning techniques have also been applied to learn from human feedback (RLHF).
Similar to the discussion in \Secref{sec:related_work}, these works typically learn a \textit{sequence-level} classifier that predicts human (pairwise) preferences and during LM training add a general-purpose KL penalty that is less-targeted to the specific LM task and feedback (preference, metric scores, \etc) \citep[\eg,][]{instructgpt2022,ziegler2019fine,bai2022training,menick2022teaching}, such as a token-level KL penalty towards the initial LM prior to training.

Alternatively, the divergence of the LMs from a target distribution can also be used as the finetuning objectives.
This line of research \citep[\eg,][]{khalifa2021a,korbak2022reinforcement,go2023aligning} formalizes controlled text generation as a constraint satisfaction problem over LM's probability distribution, with an additional divergence-minimization objective that the LMs should have a minimal KL- or $f$-divergence from the original pretrained model.
These approaches, however, require explicit functional specification on the constraints  or on the human preference, rather a more vague form of (binary) comparison between LM samples. 
For example, \citet{go2023aligning} consider human preference as a probability distribution measuring how well the generated text-sequence satisfies the preference.
Apart from this more demanding requirement, these approaches further require special methods to sample from the resulting LM.

To sum up, prior works on aligning LMs with preference mostly focus on an ungrounded \textit{sequence-level}  guidance, which can suffer from the delay-feedback issue in LM training, as discussed in Sections \ref{sec:intro} and \ref{sec:related_work}.
By contrast, our preference-grounding perspective can provide a stable, data-driven, task-specific \textit{token-level} guidance on LM training, and can potentially improve on the vanilla MLE, especially when the quality of the supervised data cannot be guaranteed.
We experimentally validate this intuition in \Secref{sec:exp} and Appendix~\ref{sec:more_abla}.

Apart from fine-tuning the pretrained LMs, \citet{korbak2023pretraining} recently apply preference alignment to the pre-training stage of the LMs.
As with prior works, the sparse sequence-level evaluation (without KL penalty/stabilizer) is directly used, to learn a token-level value function, to condition the LM generation on, or for a reward-weighted regression objective. 
The pre-training stage in \citet{korbak2023pretraining} is a potential further application of our framework since we make no assumption on the zero-shot ability of the initialized LMs, as discussed in Sections \ref{sec:policy_training} and \ref{sec:exp_abla}.

We also notice that a recent robotics paper \citep{kim2023preference} proposes to \textit{learn} a \textit{weighted-sum} aggregation  together with the per-step reward, to form the sequence-level evaluation in learning the reward function, based on pairwise preference over two trajectories of equal length.
Compared with this recent work, our aggregation functions in \Secref{sec:method_agg_func} do not require additional modeling and training, and therefore can be more efficient and more stable for the reward-function learning.
Additionally, we do not assume that trajectory lengths are equal, as this may be infeasible for LM tasks such as text summarization.
Furthermore, our framework allows utilizing the preference among more than two trajectories, rather than the classical pairwise preference. 
In this particular aspect, our framework can be more general than this recent work of \citet{kim2023preference}.

\section{A Discussion on Applying RL Methods to LM Tasks} \label{sec:details_rl_nlp}
\subsection{LM Generation as a Token-level MDP} \label{sec:details_lm_token_mdp}

In most LM generation tasks, there is a dataset $\gD = \{(x^{i}, y^{i})\}_{i=1}^N$ of $N$ supervised examples, where $x$ is the input to the LM that can be a dummy, and $y \in \gY$ is the target text sequence.
Viewing the LM as a token-level RL policy, LM generation can be formulated as a sequential decision-making problem, specified by the Markov Decision Process (MDP) $\gM = (\sS, \sA, P, \gR, \gamma, \mu_0)$ \citep{rlintro2018}.  
Specifically, $\sS$ is the state space, where the state at timestep $t$, $s_t$, consists of the LM input $x$ and the previously generated tokens $a_{< t} = (a_0, \ldots, a_{t-1}), t > 0$, \ie, $s_0 = x$ and $\forall\, t > 0, s_t = (x, a_{< t})$. 
$\sA$ is the action space, which is the vocabulary $\gV$, and an action $a_t$ at timestep $t\geq 0$ is a token from~$\gV$.
$P(s_t, a_t): \sS \times \sA \rightarrow \sS$ is the transition function that deterministically appends the newly sampled token to the end of the current state, \ie, $\forall\, t \geq 0, s_{t+1} = (s_t, a_t) = (x, a_{\leq t})$.
$\gR(s_T, y): \sS \times \gY \rightarrow \R$ is the environmental reward (task-specific evaluation metric) that depends on the \textit{final state} $s_T$ of the LM-generation trajectory and the target sequence $y$.
Here $T$ is the ending time of the trajectory, \ie, the length of the full generated  text sequence; and $s_T = (x, a_0, \ldots, a_{T-1})$ is the final state of the generation trajectory consisting of the LM input $x$ and the full generated text sequence $\va = (a_0, \ldots, a_{T-1})$.
$\gamma \in [0,1]$ is the discount factor.
And $\mu_0(x): \sS \rightarrow [0,1]$ is the distribution of the initial input $x$.

We denote the LM as $\pi_\theta(a_t \given s_t)$, parametrized by~$\theta$.
At each timestep $t$, $\pi_\theta(a_t \given s_t)$ generates the next token $a_t$ given the current state $s_t = (x, a_{< t})$.
The ultimate goal of policy learning (LM training) is to maximize the expected environmental reward $\gR$, which can be expressed as
\begin{equation*}
\textstyle
        \max_\theta \E_{(x,y)}\E_{\va \sim \prod_{t=0}^{T-1} \pi_\theta(a_t \given s_t)}\sbr{\gR(s_T = (x, \va), y)}\,,
\end{equation*}
where $(x,y)$ is drawn from the corresponding sampling distribution.


\subsection{Delayed Feedback in RL-based LM Training}\label{sec:delayed_feedback_rl_lm}
As discussed in Appendix~\ref{sec:details_lm_token_mdp}, the environmental reward $\gR(s_T, y)$ is only defined on the full generated text sequence $\va$. 
The token-level MDP formulation of LM generation thus
meets the problem of sparse reward-signal or the delayed feedback issue discussed in \Secref{sec:intro}.
Hereafter, we will use ``sparse reward (signal)'' and ``delayed feedback'' interchangeably depending on the context, as they are used synonymously in the RL literature.

Specifically, prior works \citep[\eg,][]{sqltext2021,rlprompt2022,snell2022offline} often manually interpolate the intermediate rewards by some non-informative values such as $0$ or $-1$, \ie, $\forall\, t \geq 0$
\begin{equation} \label{eq:sparse_reward}
    \gR(s_t, y) = \begin{cases}
        0 \text{ or } -1, \quad t < T\\
        \gR(s_T, y),\quad t = T
    \end{cases}.
\end{equation}
It is clear that the reward signal is sparse. 
In other words, the feedback to intermediate actions/tokens is delayed until the full text-sequence has been generated.

We note that this sparse-reward/delayed-feedback problem will not be addressed by the standard actor-critic or Q-learning methods in RL.
With only sparse reward-signals, it can be difficult to estimate the token-level value functions in these RL methods.

Specifically, the standard Monte Carlo estimate of the value functions is known to have high variance due to the large sampling space \citep{rlintro2018}. 
This problem is even severe in the LM tasks where there are exponentially many text sequences that can follow a partial sequence.

Further, as discussed in \citet{sqltext2021}, the sparse-reward/delayed-feedback problem can also hurt the bootstrapping-style method for learning the value functions, since
the standard value-function learning can suffer from ``the unstable per-step bootstrapping-style training with sparse reward signals.''
This can subsequently harm the LM training since many actor-critic or Q-learning methods rely heavily on how accurately the learned value function(s) can assess the quality of intermediate text sequences \citep{sqltext2021,rlintro2018,jointmatching2022}.

\subsection{Sparse Reward with KL Penalty}

With the sparse-reward/delayed-feedback issue in Appendix~\ref{sec:delayed_feedback_rl_lm}, prior works typically add a token-level KL-penalty to the sparse sequence-level environmental rewards Eq.~\eqref{eq:sparse_reward}.
For simplicity, assume that in Eq.~\eqref{eq:sparse_reward} the intermediate rewards are interpolated by $0$.
The KL-stabilized reward signal $R(s_t, a_t, y)$ is
\begin{equation} 
    R(s_t, a_t, y) = \begin{cases}
        - c \cdot \mathrm{KL}(\pi_\theta(a_t \given s_t) \,||\, \pi_0(a_t \given s_t)), & t < T-1\\
        \gR(s_T, y) - c \cdot \mathrm{KL}(\pi_\theta(a_t \given s_t) \,||\, \pi_0(a_t \given s_t)), & t = T - 1
    \end{cases},
\end{equation}
where $c$ is a hyper-parameter and $\pi_0$ is some prior distribution, such as the uniform distribution \citep{sqltext2021,rlprompt2022}, the initial LM prior to training \citep{ziegler2019fine,nlpo2022}, the supervised-fine-tuned model \citep{offlinerldialog2019,jaques2020human,stiennon2020learning,instructgpt2022}, or the base momentum model \citep{castricato2022robust}.
For a concrete example, see Line 224-235 of the popular \href{https://github.com/CarperAI/trlx/blob/0c5246f64e5e0ecb5fb2de65d440b122c792caf8/trlx/orchestrator/ppo_orchestrator.py#L224}{trlx package's implementation}.

With this KL-stabilized reward signal $R(s_t, a_t, y)$, the action-value function for the policy/LM $\pi_\theta$ is 
\begin{equation} \label{eq:q_func_kl_reward}
\resizebox{.94\linewidth}{!}{%
$
     \begin{aligned}
    Q(s_t, a_t, y) &= \E_{\cbr{a_{t'}}_{t'=t+1}^{T-1} \sim \pi_\theta} \sbr{\sum_{t'=t}^{T-1} \gamma^{t'-t} R(s_{t'}, a_{t'}, y) \given s_t, a_t} \\
    &= \E_{\cbr{a_{t'}}_{t'=t+1}^{T-1} \sim \pi_\theta} \sbr{\gamma^{T-1-t}\gR(s_T, y) -c \cdot \sum_{t'=t}^{T-1} \gamma^{t'-t} \mathrm{KL}(\pi_\theta(a_{t'} \given s_{t'}) \,||\, \pi_0(a_{t'} \given s_{t'})) \given s_t, a_t}
\end{aligned}
$%
}
\end{equation}
It is clear from Eq.~\eqref{eq:q_func_kl_reward} that the environmental reward $\gR(s_T, y)$ is multiplied by a factor exponentially decayed with respect to the length of the remaining horizon $T-1-t$.
Without the KL penalty, the action-value $Q(s_t, a_t, y)$ could be tiny when $t$ is small, \ie, at the beginning of the text-sequence generation.
This could make it hard to accurately model and learn the action values, echoing 
the previously-stated harm of the sparse-reward/delayed-feedback problem mentioned by \citet{sqltext2021}

Recall that the standard actor-critic and Q-learning methods in RL use the action-value function $Q(s_t, a_t, y)$ as the token-level guidance (per-step critic) for policy/LM training.
Due to the exponentially decaying factor $\gamma^{T-1-t}$, 
when the discount factor $\gamma$ in Eq.~\eqref{eq:q_func_kl_reward} is not sufficiently large, this token-level guidance $Q(s_t, a_t, y)$ in RL-based LM training mainly reflects the (discounted) sum of future KL-penalty, rather than the actual goal of LM training --- the  environmental reward $\gR(s_T, y)$. 
This phenomenon can be more evident at the beginning of the text-sequence generation, \ie, when the length of the remaining horizon $T-1-t$ is long.
On the other hand, learning the action-value function $Q(s_t, a_t, y)$ under a large discount factor $\gamma$ is known to be challenging \citep{rlintro2018}, since the highly varying (late) future can significantly affect the current action value $Q(s_t, a_t, y)$.
The selection of the discount factor $\gamma$, therefore, becomes a tradeoff and a challenge.
Note that $\gR(s_T, y)$ here is generic and can represent automatic evaluation metrics or (human) preference, and that the beginning of text generation can affect all subsequent token selections.
Intuitively, using Eq.~\eqref{eq:q_func_kl_reward} as the token-level guidance for policy/LM training can thus be less successful in the concrete LM task, especially when generating longer sequences, as we verified in Appendix~\ref{sec:more_abla}.

In the experiments (\Secref{sec:exp} and Appendix~\ref{sec:more_abla}), we compare our preference-grounding approach with RL-based baselines that estimate a standard value function similar to Eq.~\eqref{eq:q_func_kl_reward} from sparse environmental reward with KL penalty, such as the RLPrompt method \citep{rlprompt2022} and the (supervised+) PPO/NLPO methods in RL4LMs \citep{nlpo2022}.
We leave as future work the potential combination of our preference-grounded guidance with actor-critic and Q-learning methods in RL-based LM training.

\section{Further Discussion on the Guidance Re-estimation Scheme} \label{sec:details_rew_retrain_scheme}

As discussed in \Secref{sec:policy_training}, in this paper, we deal with the most general setting where the LM training directly starts from a raw pre-trained LM, rather than an initial LM that has been fine-tuned via supervised learning on the desired dataset, such as in \citet{stiennon2020learning}. 
We also make no assumptions about the zero-shot ability of the raw pre-trained LM. 
We choose this setting because it is more general and naturally fits into the task of text-prompt generation, where supervised datasets of good prompts are not available and the initial LM cannot generate good prompts.

As discussed before, under this general setting, the LM $\pi_\theta$ can evolve from a less-preferred distribution to a highly-preferred one, over the training process. 
Since our reward function $r_\phi$ is trained by text sequences sampled from $\pi_\theta$, there is a distribution shift between the sequences used to train $r_\phi$ during reward-function learning, and the sequences evaluated by $r_\phi$ during LM training, especially after $\pi_\theta$ has been sufficiently improved. 
To keep $r_\phi$ as accurate guidance for LM training, a natural idea is to refine $r_\phi$ periodically on the text generations from the latest LM, leading to our reward-function retraining scheme. 

We emphasize that \textit{the reward-function retraining scheme does not give our method an unfair advantage over the baseline methods}. 
In particular, RLPrompt \citep{rlprompt2022} and RL4LMs’ methods \citep{nlpo2022} retrain their value-functions in every optimization step, and thus, they query the environmental reward in every optimization step. 
Specifically, in Algorithm 1 of the RL4LMs paper, the penalized reward $\hat R_t$ is calculated in each optimization step, whose calculation requires the true environmental reward $R$ (Eq. (1) of the RL4LMs paper). 
Besides, in the codebase of RLPrompt, this environmental interaction is implemented in  \href{https://github.com/mingkaid/rl-prompt/blob/24ff3e6a81bbd39e4d9ccaaaee41885bc5058682/rlprompt/modules/sql_module.py#L125}{this line}, which is queried in every optimization step, as seen in \href{https://github.com/mingkaid/rl-prompt/blob/24ff3e6a81bbd39e4d9ccaaaee41885bc5058682/rlprompt/trainers/trainer.py#L158}{this line}.
In the notion of Reinforcement Learning from Human Feedback (RLHF), this every-step interaction is similar to asking humans to score the LM generations in every training step, which can be infeasible. 
By contrast, in our paper, we reduce the frequency of these environmental interactions by retraining the guidance model only periodically and only during the first half of the LM-training process.

Though the motivation of this reward-function  retraining scheme comes from model-based RL (\Secref{sec:policy_training}), we notice that some prior RLHF works do implement similar ideas. 
For example, Page~2 of \citet{ziegler2019fine} mentions that ``..., we continue to collect additional data and retrain our reward model as the policy improves (online data collection).''
Page 2 of \citet{stiennon2020learning} mentions that ``We can then gather more human data using samples from the resulting policy, and repeat the process.'' 
Page 5 of \citet{menick2022teaching} and Page 20 of \citet{bai2022training} also have similar discussions.
Based on these, our reward-function retraining scheme is both well-motivated and practical, even with human rankings in RLHF.

\section{Potential Negative Societal Impacts}



Since our framework can ground the sequence-level preference into token-level guidance for LM training and can be not tied to a specific preference source, it is possible that this framework may be used to train ill-intended LMs by grounding some malicious or unethical preferences.
This potential negative impact may be mitigated by closer monitoring the datasets on which our framework operates.

\section{Limitations}

Since our token-level guidance is learned by grounding sequence-level preference,
a potential failure case of our framework will be when the preference orderings are very noisy. In this situation, the learned guidance may not be meaningful and hence could even deteriorate the subsequent utilization of it in LM training.

Even though we have shown in \Secref{sec:exp_abla} that it can be beneficial to use more than two sequences to learn the token-level guidance, it can be practically challenging to obtain a high-quality ranking among many candidate text sequences, \eg, when the number of sequences is more than seven.

Besides, the reward-function retraining scheme may incur some additional computational complexity, compared with training the reward function only once and fixing it throughout the LM-training process.

\section{Computational Resources}
The experiments are conducted on NVIDIA GeForce RTX 3090 and NVIDIA A100 GPUs.
Depending on the specific task and setting, several models could be trained concurrently on a single GPU.

\end{document}